\documentclass{article}

    \PassOptionsToPackage{numbers, sort&compress}{natbib}

 \usepackage[main, final]{neurips_2026}

\usepackage[utf8]{inputenc} 
\usepackage[T1]{fontenc}    
\usepackage{hyperref}       
\usepackage{url}            
\usepackage{booktabs}       
\usepackage{amsfonts}       
\usepackage{nicefrac}       
\usepackage{microtype}      
\usepackage{xcolor}         
\usepackage{graphicx}
\usepackage{multirow}
\usepackage{subcaption}

\usepackage{amsthm}

\theoremstyle{plain}      
\newtheorem{theorem}{Theorem}[section]
\newtheorem{proposition}[theorem]{Proposition}

\theoremstyle{definition} 
\newtheorem{definition}[theorem]{Definition}

\theoremstyle{remark}     

\title{SIEVE: Overcoming Topological Obstruction in Equivariant Self-Supervised Learning}

\author{%
  Jongann Lee \thanks{Equal contribution.}\\
  SolverX, UIUC\\
  Urbana, Illinois, USA \\
  \texttt{jongann2@illinois.edu} \\
  \And
  Sun Woo Park \footnotemark[1] \\
  Max Planck Institute of Mathematics \\
  Bonn, Germany \\
  \texttt{s.park@mpim-bonn.mpg.de} \\
  \And
Yun Young Choi\kern0.4em%
\thanks{Corresponding author: \texttt{young@solverx.ai}.}\\
  SolverX \\
  Seoul, Republic of Korea \\
  \texttt{young@solverx.ai} \\
}

\begin{document}

\maketitle

\begin{abstract}
    We identify a \textbf{theoretical incompatibility} between 3D point cloud masked autoencoding and SE(3)-equivariant networks. We prove that standard masking triggers a \emph{topological collapse}, effectively forcing the model to hallucinate orientation without a reference frame. To resolve this, we introduce \textbf{SIEVE} (Scalar Invariant Extraction Via Equivariance), an SE(3)-equivariant self-supervised framework that compresses point cloud geometry into invariant scalar embeddings. At its core is the \emph{spherical projection mask}, which preserves global pose while obfuscating local semantics, sidestepping the obstruction by construction. Experiments confirm that SIEVE circumvents the predicted collapse where constant and noise masks fail. On downstream point cloud tasks, the resulting embeddings improve over coordinate-based baselines and hand-crafted geometric descriptors, with the largest gains on generalization to unseen shape categories. Code is available at \url{https://github.com/voltwin-dev/SIEVE_NeurIPS26}
\end{abstract}

\section{Introduction}

Point clouds serve as the fundamental data structure for 3D understanding in various domains, including robotics \citep{gu2023maniskill2} and autonomous driving \citep{caesar2020nuscenes}. Unlike 2D images, where \emph{position} (the pixel grid) and \emph{content} (RGB values) are naturally disentangled, point clouds present a unique challenge: the geometry \emph{is} the feature. The coordinate $\mathbf x \in \mathbb R^3$ encodes both the local structure and the global spatial location. This entanglement necessitates representations that are not only topologically rich, a goal also pursued in graph representation learning \citep{choi2022cycle,choi2024topology}, but also consistent
under global rigid-body transformations.

This separation naturally motivates the use of SE(3)-equivariant neural networks, whose architectures are constrained to ensure that a rigid-body transformation applied to the input results in an identical transformation of the output \citep{satorras2021EGNN, brandstetter2021SEGNN,fuchs2020se3transformers, GATr2023, bekkers2023PONITA}. However, because these models cannot rely on absolute coordinate ``shortcuts'' \citep{wang2023extrinsic_equivariance, wu2025SONATA}, they often face more difficult optimization landscapes than their non-equivariant counterparts \citep{xie2025equivariantLoss}. We propose self-supervised learning (SSL) as the path forward: the equivariant pipeline can distill geometric structure into SE(3)-invariant embeddings that enrich downstream representations, sidestepping the need to impose equivariance on the downstream model.

However, we identify a fundamental theoretical flaw when applying standard masking strategies to SE(3)-equivariant neural networks. Conventional SSL methods replace masked features with a learnable, isotropic token \citep{he2022MAE,yu2021pointbert,pang2022pointMAE}. Utilizing the theory of pullbacks of vector bundles, we demonstrate that the use of such constant tokens induces a topological obstruction, leading to a collapse in the vector pipeline that inhibits the reconstruction of the input point cloud. We empirically validate this obstruction through a controlled experiment on a synthetic dataset, highlighting the catastrophic failure of isotropic masking in equivariant settings. 

To resolve this, we propose the equivariant SSL framework \textbf{SIEVE} (Scalar Invariant Extraction Via Equivariance). Instead of masking information completely, SIEVE employs a \emph{spherical projection masking strategy} that \emph{sieves} the information: it projects the `where' (global pose) onto the sphere while forcing the encoder to compress the `what' (intrinsic semantics) into invariant scalars. This masking scheme is a direct consequence of our vector bundle analysis: it materializes the pullback of non-trivial vector bundles that standard masking cannot reconstruct. We validate these embeddings on several downstream point cloud tasks, where SIEVE improves over coordinate-based baselines and hand-crafted descriptors with the largest gains appearing on generalization to unseen categories.

\section{Theoretical Framework} \label{sec:Theory}

\begin{figure}[t]
    \begin{center}\centerline{\includegraphics[width=0.9\textwidth]{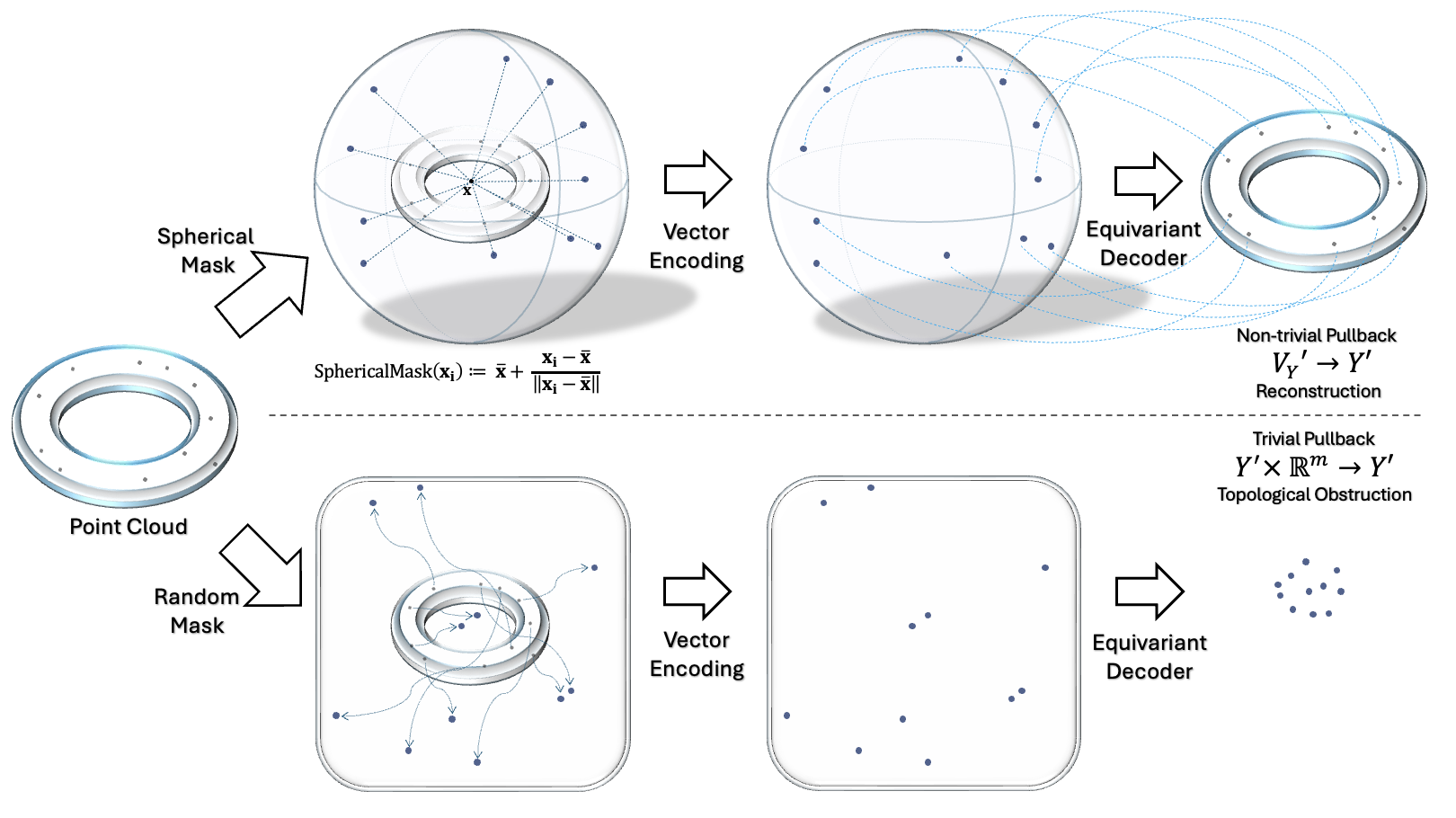}}
        \vskip -0.1in
        \caption{\textbf{Random masking causes representational collapse; spherical projection masking does not.} Random masking functions ignore subtle geometric data inherent in point clouds. This imposes topological obstructions in reconstructing the desired non-trivial vector bundles characterizing the underlying geometry of point clouds, as demonstrated in Theorem \ref{thm:constant_obstruction}. As a result, representational collapse can occur as shown in Section \ref{subsec:MaskAblation}. SIEVE utilizes a spherical projection mask, thereby possessing capabilities in reconstructing a class of pullbacks of vector bundles defined over a wide variety of 2 and 3-dimensional manifolds as demonstrated in Theorem \ref{thm:SIEVE_main}.}
        \label{fig:SIEVE_Comparison}
    \end{center}
    \vskip -0.2in
\end{figure}

\subsection{Preliminaries: Vector Bundles}\label{subsec:Theory_preliminaries}

Effective analysis of properties of 3D point clouds rests upon understanding entanglement between \textit{position} and \textit{content}. For this purpose, we seek suitable mathematical concepts or frameworks that encompass both topological features of 3D point clouds as well as contents measured over such point clouds as real vectors. We pay particular focus on using vector bundles from algebraic topology among many possible candidates that can harbor the subtlety of analyzing such entanglements.

\begin{definition}
    \label{def:vector_bundle}
    Let $X$ be a topological space. An $n$-dimensional vector bundle $\pi: V \to X$ is a topological space satisfying the following conditions. \textbf{(1)} The map $\pi$ is a continuous surjection. \textbf{(2)} For every $x \in X$, the fiber $\pi^{-1}(x)$ is isomorphic to $\mathbb{R}^n$. \textbf{(3)} For every $x \in X$, there exists an open neighborhood $U \subset X$ such that satisfies the following three conditions. \textbf{(3a)} There exists a homeomorphism $\varphi: U \times \mathbb{R}^n \to \pi^{-1}(U)$. \textbf{(3b)} For every $v \in \mathbb{R}^n$, we have $\pi(\varphi(x,v)) = x$. \textbf{(3c)} Every isomorphism $\mathbb{R}^n \to \pi^{-1}(x)$ can be rewritten as an invertible $n \times n$ real matrix.
\end{definition}

Interested readers may refer to Appendix \ref{appendix:motivation} for a detailed motivation for using the notion of vector bundles to formulate strengths and capabilities of deep learning algorithms. Given a vector bundle $\pi: V \to X$, a continuous function $f: X \to V$ such that $\pi(f(x)) = x$ for every $x \in X$ is called a global section. With these definitions, we propose the following topological reformulation of 3D point clouds, up to some perturbations or negligible errors. \textbf{(1)} \textit{3D point clouds} correspond to points on a smooth compact orientable 3-dimensional \textit{manifold} $Y$ embedded inside $\mathbb{R}^3$, whose boundary $X := \partial Y$ is a smooth compact orientable surface embedded inside $\mathbb{R}^3$. 
\textbf{(2)} 
Because 3D Point clouds consist of discrete set of points, we may require all the point clouds to lie over a complement of a small tubular neighborhood of the wedge sum of some circles $\vee_{i=1}^g S^1$ inside $Y$, where $g$ is the genus of $X$. Such a space, denoted as $Y'$, is homotopy equivalent to $X \times I$. \textbf{(3)} \textit{Measurements} consisting of $n$-dimensional feature vectors over each point cloud $x \in Y'$ corresponds to \textit{global sections} of an $n$-dimensional vector bundle $V \to Y'$.

It is natural to understand vector bundles that can harbor non-trivial entanglements between geometry and intrinsic semantics which 3D point clouds contain. This brings us to the notion of trivial vector bundles over topological spaces.
\begin{definition}
    Let $X$ be a topological space. An $n$-dimensional trivial vector bundle $V \to X$ is a vector bundle over $X$ which is isomorphic to the vector bundle $X \times \mathbb{R}^n \to X$. Any continuous function $f: X \to \mathbb{R}^n$ is a global section of the trivial vector bundle.
\end{definition}
The geometry of $X$ does not impose any restrictions on the image of the global section $f: X \to \mathbb{R}^n$. We hence obtain the following insight: \textbf{trivial vector bundles} over topological spaces are suitable for analyzing data whose \textit{position} and \textit{content} are \textbf{disentangled}. We also observe that vector bundles over topological spaces which are \textbf{non-trivial} are potential candidates that can illuminate \textbf{entanglements} between geometry and intrinsic semantics of 3D point clouds. In light of these insights, we ask the following question: \textit{What are effective ways to learn global sections of real oriented non-trivial vector bundles over a given topological space $Y$ (or $Y'$)?}

\subsection{Interpretation: Simplifications with Projections to $S^2$ using Pullback of Vector Bundles} \label{subsection:mathematical-interpretation}

To understand the question, we analyze vector bundles over $Y$ (or $Y'$) that may be underlying spaces for 3D point clouds. The problem of classifying real oriented vector bundles over smooth manifolds, however, is a highly non-trivial problem. Nevertheless, we can construct some of these desired vector bundles by using pullbacks of vector bundles between topological spaces.
\begin{definition}
  \label{def:pullback}
    Given a continuous map $f: X \to Y$, and a vector bundle $\pi: V \to Y$, the pullback of $V$ is the vector bundle $f^*\pi: f^*V \to X$, where $f^*V := \left\{ (x,e) \in X \times V : f(x) = \pi(e) \right\}.$
    The fiber of $f^*V$ at $x \in X$ is isomorphic to the fiber of $V$ at $f(x) \in Y$.
\end{definition}
Without loss of generality, we consider an auxiliary (possibly unknown) $n$-dimensional vector bundle $V_Y \to Y'$ underlying 3D point clouds, where $n \geq 3$. Because $Y'$ and $X$ are homotopy equivalent, there exists a bijection of isomorphism classes of vector bundles of $Y'$ and $X$ (see Corollary 1.8 of~\citep{HatcherVectorBundle}). Hence, we can identify the $n$-dimensional vector bundle $V_Y \to Y'$ with an $n$-dimensional vector bundle $V_X \to X$. (This is one of the reasons why we chose $Y'$ instead of $Y$, because we may reduce the dimension of the underlying topological space of our interest from $3$ to $2$.) 

We can use pullbacks of vector bundles to construct trivial and non-trivial vector bundles over $X$. Given any continuous map $f: X \to S^2$, we can pullback some $n$-dimensional vector bundle $V \to S^2$ to construct a vector bundle $f^*V \to X$. Thankfully, the classification of real oriented vector bundles over $S^2$ is very well understood, whose proof is in Appendix~\ref{appendix:proof-main_theorem}. 
\begin{theorem}
  \label{thm:main_theorem}
  Let $V \to S^2$ be a real oriented vector bundle of dimension $n \geq 3$. Suppose $V$ is a non-trivial vector bundle, i.e. $V$ contains the tautological bundle $\pi_t: T \to S^2$ as a sub-vector bundle (see Appendix~\ref{appendix:proof-main_theorem} for its definition). Then $V$ is isomorphic to the \textbf{unique} non-trivial vector bundle of dimension $n$, whose fibers are isomorphic to $\mathbb{R}^{n-2} \oplus \pi_t^{-1}(x)$ for every $x \in S^2$.
\end{theorem}
There are two isomorphism classes of vector bundles over $X$ we can construct using pullback of vector bundles of $S^2$: the trivial vector bundle $X \times \mathbb{R}^n \to X$, obtained as a pullback of the trivial vector bundle $S^2 \times \mathbb{R}^n \to S^2$, and a non-trivial vector bundle $V_X \to X$, obtained as a pullback of the unique non-trivial vector bundle $V \to S^2$.

\subsection{Applications: Topological Obstructions in Constant and Randomized Noise Masking}

The main application of our paper will focus on effectively approximating 3D point clouds, which shall be regarded as a set of discrete points $\mathcal{X}$ over $Y'$. Given an injection $i: \mathcal{X} \hookrightarrow Y'$, we consider the restriction of the global section $f \circ i: \mathcal{X} \to Y' \to V_Y'$. We regard the problem of approximating 3D point clouds as a problem of approximating the restriction of the global section $f \circ i$. 

To make the notion of approximation $f \circ i$ rigorous, it suffices to clarify what it means for an $\mathrm{SE}(3)$-equivariant neural network $\Psi$ to effectively approximate a global section $f: Y' \to V_Y'$. To do this, we identify a global section $f: Y' \to V_{Y}'$ of an $N$-dimensional vector bundle $V_Y'$ over a $3$-dimensional manifold $Y'$ constructed in Section \ref{subsection:mathematical-interpretation} as a restriction of the continuous function $F: \mathbb{R}^3 \to \mathbb{R}^{m_N}$ to $Y'$ (or any connected compact subspace of $\mathbb{R}^3$ homeomorphic to $Y'$) that is equivariant under the action of $\mathrm{SE}(3)$. Here, the quantity $m_N$ may not necessarily be equal to $N$, but can be defined as the minimal dimension $m_N \geq N$ of a trivial vector bundle $\mathcal{T}_Y'$ over $Y'$ which contains $V_{Y}'$ as a vector sub-bundle. We note that this quantity is guaranteed to exist using Proposition \ref{appendix:prop:trivial_bundle}, whose statement appears in the next subsection. More concretely, the relation between $f$ and $F$ is given as follows. Given an inclusion $j: Y' \to \mathbb{R}^3$ and a projection $\pi: \mathcal{T}_Y' \to V_Y'$, we have
\begin{equation}
    f = \pi \circ F \circ j: Y' \to \mathbb{R}^3 \to \mathbb{R}^{m_N} \to \mathbb{R}^N.
\end{equation}
Using the analogous argument, we note that the restriction of a global section $f \circ i$ over a 3D point cloud $\mathcal{X}$ can be obtained from a continuous function $F: \mathbb{R}^3 \to \mathbb{R}^{m_N}$ via the composition $\pi \circ F \circ j \circ i$.

Based on this identification, we define what it means for an $\mathrm{SE}(3)$-equivariant neural network to effectively approximate a global section. Similar definitions can be found in Theorem 11 of \citep{2021KS} and Theorem 4.1 of \citep{2020PV}.
\begin{definition} \label{defn:effective_approximation}
    Let $\Psi$ be an $\mathrm{SE}(3)$-equivariant neural network whose activation functions are non-constant and Lipschitz continuous. We say that $\Psi$ effectively approximates a global section $f:Y' \to V_Y'$ if for any $\epsilon > 0$ the neural network $\Psi$ satisfies the following inequality:
    \begin{equation}
        \sup_{z \in Y' \subset \mathbb{R}^3} |\Psi(z) - F(z)| < \epsilon.
    \end{equation}
    Given a discrete point cloud $\mathcal{X}$ on $Y'$ endowed with an injection $i: \mathcal{X} \to Y'$, we say that $\Psi$ effectively approximates a restriction of a global section $f \circ i:\mathcal{X} \to V_Y'$ if for any $\epsilon > 0$ the neural network $\Psi$ satisfies the following inequality:
    \begin{equation}
        \sup_{z \in \mathcal{X} \subset \mathbb{R}^3} |\Psi(z) - F(z)| < \epsilon.
    \end{equation}
\end{definition}
By formulating the notion of effective approximation using continuous functions over manifolds, instead of discrete point clouds, we can apply Definition \ref{defn:effective_approximation} to any 3D point cloud $\mathcal{X}$ on $Y'$ once we specify an injection $i: \mathcal{X} \to Y'$. By doing so, one can hope to ensure robustness in approximating the desired restriction $f \circ i$ given any arbitrary 3D point cloud $\mathcal{X}$ over $Y'$.

With the theoretic framework in place, we shall analyze potential topological obstructions imposed on certain masking functions. Given an integer $N$, let $\mathfrak{m}: X \to \mathbb{R}^N$ be a masking function that does not lift to a section of a non-trivial vector bundle $V_X \to X$. As an example, we may consider constant and randomized masking functions, which are default choices for reconstruction based SSL in both images and point clouds. We note that the image $\mathfrak{m}(\textbf{x})$ is chosen randomly independent of the geometric features of $\mathbf{x}$. We can hence interpret the application of the masking function $\mathfrak{m}$ as replacing $V_X$ with $\mathbb{R}^N \times X$. Because the trivial vector bundle over $X$ is the pullback of the trivial vector bundle over $S^2$, the masking operator by $\mathfrak{m}$ corresponds to replacing $V_X$ with $f^*(\mathbb{R}^N \times S^2)$. There is a wide class of non-trivial vector bundles over $X$ which may not admit a smooth embedding inside $f^*(\mathbb{R}^N \times S^2)$. A summary of our aforementioned findings, focusing on constant masking functions, leads to the following result, whose proof is in Appendix~\ref{appendix:proof-constant_obstruction}.
\begin{theorem}
    \label{thm:constant_obstruction}
    Let $Y'$ be the underlying $3$-dimensional smooth orientable manifold constructed from Section~\ref{subsection:mathematical-interpretation} which represents the underlying space for a given 3D point cloud. Let $\mathfrak{m}:X \to \mathbb{R}$ be any constant masking function. Then any $\mathrm{SE}(3)$-equivariant neural network with the masking function $\mathfrak{m}$ cannot effectively approximate global sections of any non-trivial vector bundles $V_Y' \to Y'$ (as in Definition \ref{defn:effective_approximation}).
\end{theorem}
Any $\mathrm{SE}(3)$-equivariant neural network with the randomized masking function $\mathfrak{m}$ cannot effectively approximate global sections of any non-trivial vector bundle of dimension at least $N$. Even for lower dimensional non-trivial vector bundles, there are many challenges that would need to be addressed to effectively approximate such global sections, see Figure \ref{fig:SIEVE_Comparison} and Appendix~\ref{appendix:proof-constant_obstruction} for more details. 

\section{SIEVE: Scalar Invariant Extraction Via Equivariance} \label{sec:SIEVE}

\begin{figure}[t]
    \begin{center}\centerline{\includegraphics[width=\linewidth]{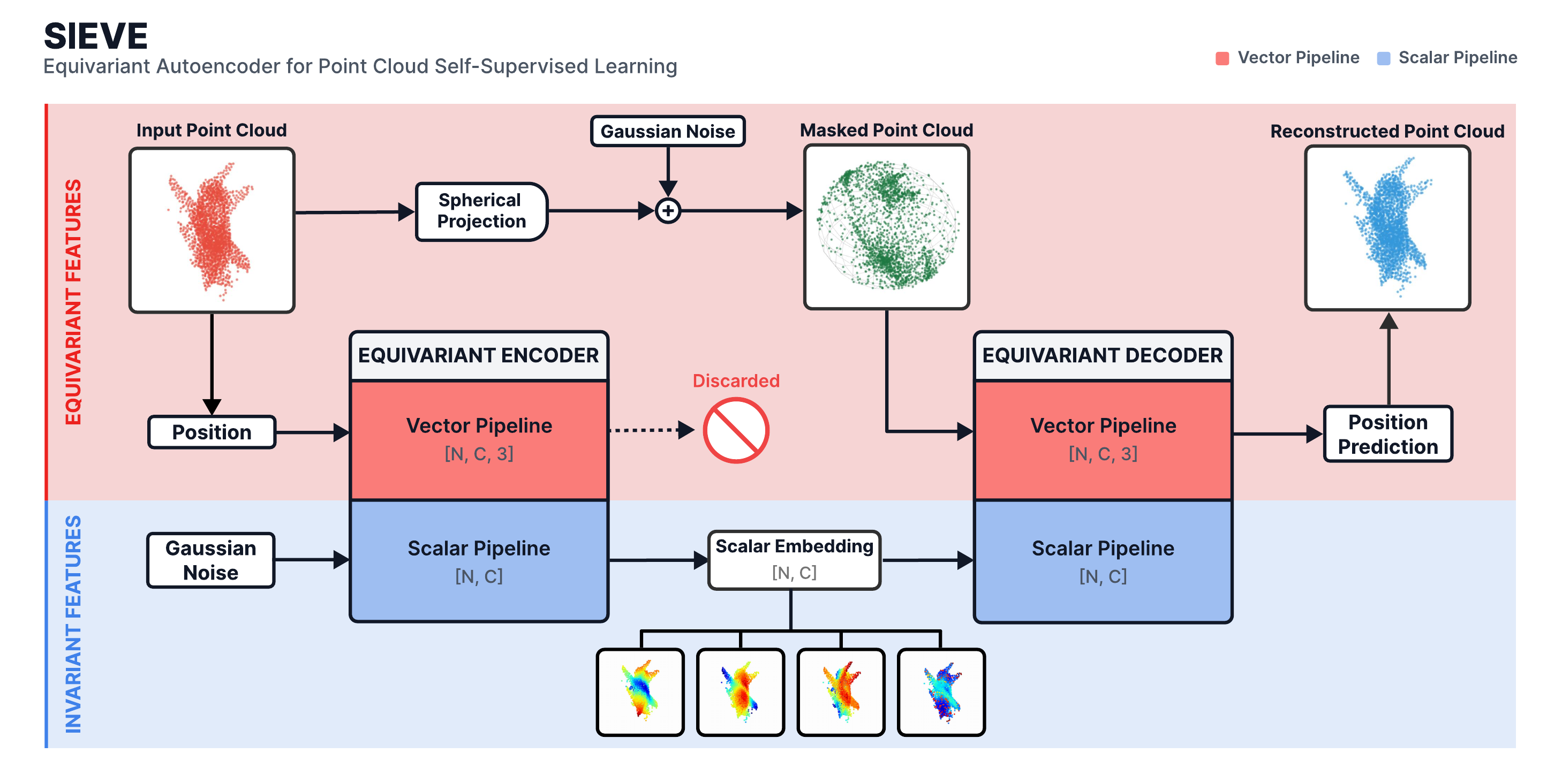}}
        \vskip -0.1in
        \caption{\textbf{SIEVE Framework.} SIEVE extracts SE(3)-invariant embeddings using self-supervised reconstruction on an equivariant bottleneck. The equivariant encoder compresses raw point coordinates into an invariant scalar encoding, while the vector pipeline is blocked. The spherical projection mask is applied to the input point cloud and supplied as the new vector embedding. It and the semantic details from the scalar embedding are used by the equivariant decoder to reconstruct the point cloud on a node-level.}
        \label{fig:SIEVE_Diagram}
    \end{center}
    \vskip -0.2in
\end{figure}

To circumvent the topological obstruction stated in Theorem \ref{thm:constant_obstruction}, we propose \textbf{SIEVE}, an SSL framework that utilizes the \emph{spherical projection mask} to maintain a continuous section in the vector bundle, thereby ensuring the representational completeness of the equivariant pipeline. SIEVE adopts a framework inspired by the Masked Autoencoder (MAE), comprising an encoder, a geometric masking operator, and a decoder \citep{he2022MAE}. The framework is based on two core design principles.


\textbf{Equivariant Feature Streams.} We leverage SE(3)-equivariant neural networks to process the data through two streams: a vector pipeline and a scalar pipeline (Figure~\ref{fig:SIEVE_Diagram}). The encoder maps the input point cloud $\mathcal X$ to an SE(3)-invariant scalar embedding $s_{code}$, capturing semantic information independent of the point cloud's global pose.

\textbf{Semantic-Geometric Decomposition.} Unlike standard MAE, which masks a subset of nodes entirely, SIEVE employs a \emph{spherical projection mask}. This operator maps all the points in $\mathcal X$ to a unit sphere, obfuscating local semantic features (the “what”) while preserving global spatial properties such as orientation and position (the “where”). This sidesteps the topological obstructions of constant or randomized vector masks, admitting a wider range of geometric entanglements for reconstruction.


\subsection{Spherical Projection Mask} \label{subsec:Spherical_Proj_Mask}

Standard masking in SSL typically replaces masked node features with a learnable constant token $\mathbf z \in \mathbb R^C$. However, applying such an isotropic mask to the vector pipeline of an SE(3)-equivariant neural network leads to a collapse of the vector output, as stated in Theorem~\ref{thm:constant_obstruction}.

To circumvent this collapse, we propose the \emph{spherical projection mask}. Given an input point cloud $\mathcal{X} = \{\mathbf{x}_i\}_{i=1}^{N_p} \subset \mathbb{R}^3$, let $\overline{\mathbf x} = \frac 1 {N_p} \sum_{i=1}^{N_p} \mathbf x_i$ denote the mean of the point cloud. The spherical projection operator projects each point $\mathbf x_i$ onto a unit sphere $S^2$ centered on the mean $\overline{\mathbf x}$:

\begin{equation}
    \mathrm{SphericalMask}(\mathbf x_i) = \overline{\mathbf x} + \frac{\mathbf x_i - \overline{\mathbf x}}{\| \mathbf x_i - \overline{\mathbf x} \|_2}.
\end{equation}
Using the mean $\overline{\mathbf x}$ as the center guarantees that projected nodes translate identically to the input nodes. Thus, the spherical projection mask is an SE(3)-equivariant operator, ensuring that SIEVE's vector pipeline is equivariant.

A naive projection leaves the scalar radius $r = \| \mathbf x_i - \overline{\mathbf x} \|$ as a trivial shortcut, leading to a representational collapse of the embeddings as demonstrated in Appendix \ref{appendix:spherical-projection-collapse}. To mitigate this, we inject i.i.d. Gaussian noise into the projected coordinates. Further mitigation strategies are discussed in Appendix \ref{appendix:pretrain-details}.

\subsection{SE(3)-Equivariant Encoder and Decoder} \label{subsec:Equi_EncDec}

The encoder and decoder in SIEVE are SE(3)-equivariant neural networks. Each layer has two feature streams: a scalar stream representing invariant features and a vector stream representing steerable features, with cross-stream interaction at every layer. We denote these feature pairs as $[s, \mathbf v]$, where $s \in \mathbb R^{N_p \times h_1}$ and $\mathbf v \in \mathbb R^{N_p \times h_2 \times 3}$.

\textbf{Encoder:} The SIEVE encoder extracts the SE(3)-invariant semantic information from the input geometry. The point cloud $\mathcal X$ is the vector input. The choice of scalar input becomes critical: we demonstrate in Appendix \ref{appendix:scalar-enc-input} that uniform scalar inputs prevent SIEVE from reconstructing the point cloud, so we provide i.i.d. Gaussian noise $s_{in} \sim \mathcal N(0, I) \in \mathbb R^{N_p}$ as the scalar feature. The encoding process is defined as

\begin{equation}
    [s_{code}, \mathbf v_{code}] = \mathrm{Encoder}([s_{in}, \mathcal X])
\end{equation}

Following the forward pass, we impose an equivariant bottleneck by discarding the vector output $\mathbf v_{code}$. The invariant scalar embedding $s_{code}$ is retained as the compressed semantic representation of the point cloud.

\textbf{Decoder:} The decoder aims to reconstruct the original point cloud $\mathcal X$ using the semantic encoding $s_{code}$ and the masked geometry $\mathbf x_{masked}$ as defined in section \ref{subsec:Spherical_Proj_Mask}. The decoder's vector output $\hat{\mathcal X}$ serves as the reconstructed point cloud, while the scalar output $s_{out}$ is discarded.

\begin{equation}
    [s_{out}, \hat{\mathcal X}] = \mathrm{Decoder}([s_{code}, \mathbf x_{masked}])
\end{equation}

\textbf{Backbone Selection:} While SIEVE is agnostic to specific SE(3)-equivariant architectures, we employ the Geometric Algebra Transformer (GATr) as our backbone \citep{GATr2023}. GATr's global attention captures long-range dependencies. It also uses memory efficient attention \citep{xFormers2022} and avoids expensive spherical harmonics, allowing for better scalability compared to other equivariant neural networks.

\subsection{Learning Objective}

The objective of SIEVE is to perform node level reconstruction, wherein each individual point $\mathbf x_i \in \mathcal X$ is mapped back to its original position. We accomplish this by using a standard $\mathcal L 1$ loss between the original point cloud $\mathcal X$ and the reconstructed point cloud $\hat{\mathcal X}$:

\begin{equation}
    \mathcal L 1 = \frac 1 {N_p} \sum_{i=1}^{N_p} \| \mathbf x_i - \hat{\mathbf x}_i \|_1.
\end{equation}


\subsection{A Mathematical Reformulation of SIEVE}

Using the theoretical motivation for SIEVE, we deduce theoretical strengths of our proposed method in comparison to other masking schemes, whose proof is in Appendix~\ref{appendix:proof_SIEVE_main}.
\begin{theorem} \label{thm:SIEVE_main}
    Let $Y'$ be the underlying $3$-dimensional smooth orientable manifold constructed from Section~\ref{subsection:mathematical-interpretation}. Then SIEVE can approximate global sections of any vector bundles which are isomorphic to pullbacks of any vector bundles of $S^2$.
\end{theorem}
The idea of the proof originates from an observation that application of $\mathrm{SphericalMask}$ corresponds to utilizing pullback of vector bundle $V \to S^2$ which contains the tautological bundle $T \to S^2$ as a vector sub-bundle. We thus deduce that SIEVE, unlike constant or random masking techniques, has theoretical capabilities in approximating all possible vector bundles over a wide range of classes of 2-dimensional or 3-dimensional closed oriented manifolds which arise from pullback of both trivial and non-trivial vector bundles over $S^2$. To support this, we present experimental demonstrations of SIEVE's theoretical capabilities in Section \ref{subsec:MaskAblation}. 

\section{Experiments}\label{sec:Experiments}

\subsection{Pre-training}\label{subsec:Pre-training}

\begin{figure}[ht]
    \vskip -0.1in
  \begin{center}
    \centerline{\includegraphics[width=0.8\linewidth]{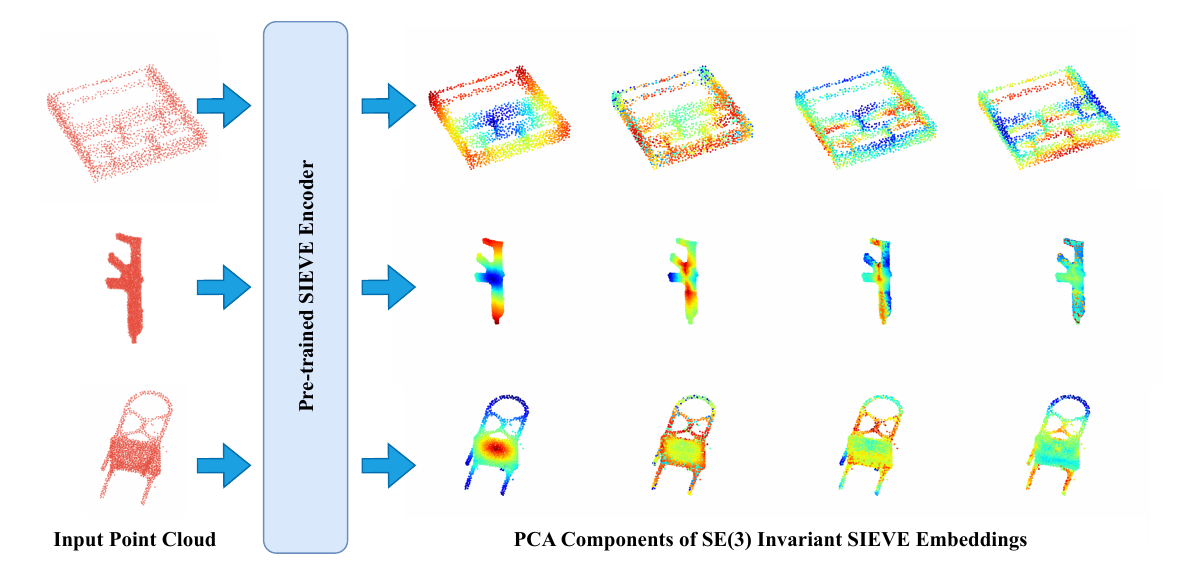}}
    \caption{
        \textbf{Pre-trained SIEVE Encoder Embeddings.} We visualize the PCA components of the pre-trained SIEVE embeddings. These embedding are invariant to point cloud rotation and translation.
    }
    \label{figure:embedding_visualization}
  \end{center}
  \vskip -0.2in
\end{figure}

We pre-train SIEVE on the OpenShape dataset, which consists of approximately 876k 3D models \citep{2023openshape}. Each point cloud is generated by sampling 2048 points via furthest point sampling. We utilize the entire dataset for pre-training without a train-validation split to maximize geometric diversity. Full pre-training details are in Appendix \ref{appendix:pretrain-details}.

\subsection{Empirical Verification of Collapse}\label{subsec:MaskAblation}

In Section \ref{sec:Theory}, we theoretically established that SE(3) equivariant networks cannot output two point clouds of different topology with a constant or noise vector token as input. To verify this, we perform an experiment using a synthetic dataset of two objects with differing topologies: a solid cuboid (homeomorphic to the 3-disk $D^3$) and a square toroid (homeomorphic to $S^1 \times S^1$). The dataset consists of 120 instances per class, partitioned into a 100/20 train-test split. Failure on a dataset this simple indicates a fundamental representational issue rather than a capacity or data limitation.


To demonstrate that this limitation is inherent to the equivariance constraint rather than a specific architecture, we evaluate two distinct backbones: GATr \citep{GATr2023} and SE(3)-Transformer \citep{fuchs2020se3transformers}. We compare three masking strategies: a constant vector mask utilizing a learnable isotropic token, an ellipsoid noise mask sampled from an ellipsoid spectrally aligned with the point cloud's principal components (and thus is SE(3)-equivariant), and our proposed spherical projection mask. All three models use identical initialization and hyperparameter settings to ensure a fair comparison.

\begin{figure}[ht]
    \vskip -0.1in
  \begin{center}
    \centerline{\includegraphics[width=0.8\linewidth]{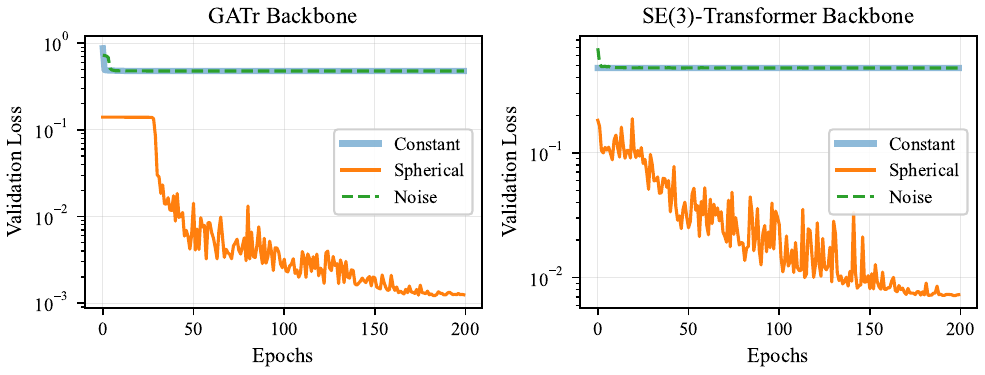}}
    \caption{
        Validation Loss of SIEVE on two baselines: GATr and SE(3)-Transformer. Trained on the synthetic dataset.
    }
    \label{figure:synthetic_loss}
  \end{center}
  \vskip -0.2in
\end{figure}


As shown in Figure \ref{figure:synthetic_loss}, models that utilize the spherical projection mask successfully converge, achieving a validation $\mathcal L 1$ loss below 0.01. Conversely, models employing the constant vector mask or the ellipsoid noise mask fail to converge, with validation and training losses stagnating around 0.5. This failure to minimize even the training loss demonstrates a fundamental representational collapse rather than a deficit in model capacity or data volume.

\begin{figure}[ht]
    \vskip -0.1in
    \begin{center}
    \centerline{\includegraphics[width=0.7\linewidth]{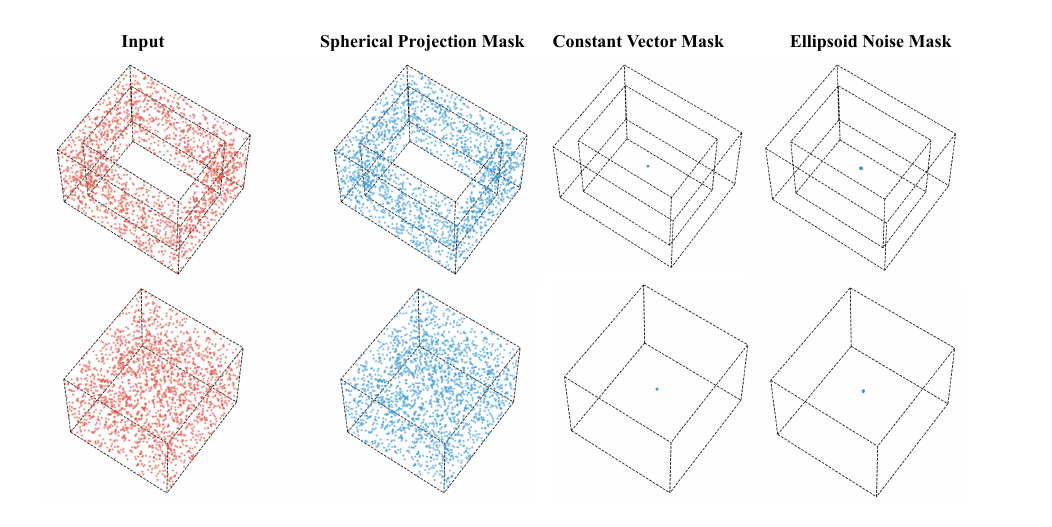}}
    \caption{
        \textbf{Reconstruction Comparison and Representational Collapse.} We evaluate the reconstruction \emph{(blue)} of the input point cloud \emph{(red)} under different masking strategies. Using our spherical projection mask, the model achieves near-perfect reconstruction. Reconstructing with a constant vector mask or the ellipsoid noise mask results in a catastrophic representational collapse. We refer to Appendix~\ref{appendix:proof-constant_obstruction} for further discussions.}
    \label{figure:synthetic_reconstruction}
  \end{center}
  \vskip -0.2in
\end{figure}

A visual inspection of the reconstructed point clouds (Figure \ref{figure:synthetic_reconstruction}) further demonstrates the predicted representational collapse of Theorem \ref{thm:constant_obstruction}. While SIEVE with the spherical projection mask achieves near-perfect reconstruction, SIEVE with the constant vector mask or the ellipsoid noise mask outputs a near zero-dimensional `blob'. As the proof of Theorem \ref{thm:constant_obstruction} suggests, these two masking schemes can only represent global sections of vector bundles of dimension at most $1$. Hence, they cannot effectively represent local coordinate functions for 3D point clouds. 

Reconstruction experiments on a dataset with identical topology confirms that it is the topological variety that causes the collapse; constant and noise mask SIEVE have no issues learning to reconstruct such datasets. (Appendix \ref{appendix:deepjeb})



\subsection{Downstream Tasks} \label{subsec:downstream_tasks}


This section evaluates whether $s_{code}$ contains useful, transferable geometric information. We do not target SOTA on any specific benchmark, as SOTA architectures are heavily tuned to their target tasks. Instead we test whether SIEVE consistently improves generalist point cloud architectures (PointNet++, DGCNN \citep{qi2017pointnet++, 2019DGCNN}). Integration is minimal: the SIEVE encoder is frozen, and $s_{code}$ is concatenated with the raw coordinates at the input layer. This drop-in usage is enabled by SIEVE's node-level granularity, which aligns one-to-one with input points, unlike patch-based SSL. We therefore evaluate on downstream tasks where such granularity is critical.

\subsubsection{Molecular Segmentation} \label{subsubsec:RNASegmentation}

We first evaluate the learned representations on the molecular segmentation task which consists of 640 RNA molecules segmented into 120 categories \citep{poulenard2019RNA_Dataset}. Following the protocol in \citep{harrison2024curvature} we resample each mesh to 2048 points and utilize the standard 80-20 train test split. 

To demonstrate that SIEVE embeddings transfer across architectures, we evaluate with three distinct architectures: PointNet++ \citep{qi2017pointnet++}, DeltaNet \citep{Wiersma2022DeltaConv}, and DiffusionNet \citep{sharp2022diffusionnet}. For each model, the frozen SIEVE scalar embedding $s_{code}$ is concatenated with the raw node coordinates $\mathbf x_i$ as input. We compare our learned representation against two SE(3)-invariant geometric representations: the distance from mean $r$ and curvature $\kappa_1, \kappa_2$. As shown in Table \ref{table:RNA_Task_Table}, SIEVE embeddings consistently outperform the ``position-only'' baselines and the geometry-enhanced models across all three architectures. 

\begin{table}[ht]
  \caption{Test accuracies(\%) on the RNA Segmentation Task}
  \label{table:RNA_Task_Table}
  \begin{center}
    \begin{small}
      \begin{sc}
        \begin{tabular}{lccccc}
          \toprule
            Model       & $xyz$  & $r$   & $\kappa_1, \kappa_2$ & \textbf{SIEVE}  \\
          \midrule
          PointNet++    & 71.55  & 75.54 & 72.90 & \textbf{86.28}  \\
          DeltaNet      & 87.21  & 88.07 & 86.86 & \textbf{89.58} \\
          DiffusionNet  & 81.20  & 85.89 & 85.97 & \textbf{89.30}  \\
          \bottomrule
        \end{tabular}
      \end{sc}
    \end{small}
  \end{center}
  \vskip -0.1in
\end{table}


\subsubsection{3D Shape Completion} \label{subsubsec:3D_EPN}

3D-EPN is a point cloud completion dataset consisting of 16 categories: 8 categories are included in both the training and test sets (``Seen''), while the remaining 8 appear only in the test set (``Unseen'') \citep{dai20173D-EPN}. We utilize PCN as our baseline and compare the performance with and without the addition of SIEVE encodings \citep{yuan2018PCN}.

As shown in Table \ref{table:3D-EPN_Table}, SIEVE consistently reduces the Chamfer Distance (CD) loss across both seen and unseen categories. The performance gain is most pronounced on unseen categories, with a $12.5\%$ reduction in the average loss compared to the $1.62\%$ reduction on seen categories. This suggests that the geometric rigor distilled into SIEVE embeddings provides a more robust structural prior than coordinate-based features alone, facilitating superior zero-shot generalization to novel topological structures.

\begin{table}[ht]
  \caption{Reconstruction $\mathcal L 2$ CD Loss $(\times 10^4)$ on the 3D-EPN Dataset}
  \label{table:3D-EPN_Table}
  \begin{center}
    \begin{small}
      \begin{sc}
        \begin{tabular}{lcclcc}
          \toprule
          Seen Categories & PCN & PCN + SIEVE & Unseen Categories & PCN & PCN + SIEVE \\
          \midrule
          Airplane    & 1.6508  & 1.8674      & Bus        & 4.2137  & 4.2908 \\
          Cabinet     & 4.8287  & 4.3237      & Bed        & 33.1502 & 29.4725 \\
          Car         & 2.6543  & 2.7663      & Bookshelf  & 13.6558 & 12.5375 \\
          Chair       & 5.7248  & 5.6236      & Bench      & 7.4325  & 6.3300 \\
          Lamp        & 7.4924  & 7.2977      & Guitar     & 7.4999  & 5.2607 \\
          Sofa        & 5.7775  & 5.6728      & Motorbike  & 9.4908  & 8.9940 \\
          Table       & 4.4875  & 4.6435      & Skateboard & 7.7232  & 6.4723 \\
          Vessel      & 4.4766  & 4.2968      & Pistol     & 12.7125 & 10.2674 \\
          \midrule
          Seen Average & 4.6366 & 4.5615      & Unseen Average & 11.9848 & 10.4869 \\
          \bottomrule
        \end{tabular}
      \end{sc}
    \end{small}
  \end{center}
  \vskip -0.1in
\end{table}

\subsubsection{Geometric Registration} \label{subsubsec:registration}

Geometric registration is the prediction of a rigid body transformation to align one shape to another. We follow the task and architecture in Deep Closest Point (DCP) \citep{wang2019dcp}. A point cloud is sampled from the ModelNet40 dataset \citep{wu2015ModelNet40}, which is then cloned, point order shuffled, then transformed with a random rigid body transformation. We embed the SIEVE embeddings to DCP baseline and test with two train-test splits: one where all 40 categories are in both sets, and the other where 20 are used in train and the other 20 are reserved for testing.

The results show that SIEVE embeddings improve both the shared and separate category registration, with more dramatic gains in the separate category. The geometric registration performance with SIEVE is much more robust to unseen objects. This pattern is consistent with the 3D-EPN results: SIEVE embeddings encode intrinsic geometric structure that transfers across categories.

\begin{table}[ht]
  \caption{Point Cloud Registration Results on the ModelNet40 Dataset}
  \label{table:DCP_Table}
  \begin{center}
    \begin{small}
      \begin{sc}
        \begin{tabular}{lcccccc}
          \toprule
          Model       & MSE($R$) & RMSE($R$) & MAE($R$) & MSE($t$) & RMSE($t$) & MAE($t$) \\
          \midrule
          \multicolumn{7}{c}{\textbf{Shared Categories}} \\
          DCP         & 1.475760 & 1.214808  & 0.826500 & 0.000004 & 0.002040  & 0.001239 \\
          DCP + SIEVE & 0.703465 & 0.838728  & 0.479402 & 0.000002 & 0.001539  & 0.000774 \\
          \midrule
          \multicolumn{7}{c}{\textbf{Separate Categories}} \\
          DCP         & 6.948212 & 2.635946  & 1.782947 & 0.000025 & 0.005048  & 0.003669 \\
          DCP + SIEVE & 0.874955 & 0.935390  & 0.511852 & 0.000002 & 0.001509  & 0.000813 \\
          \bottomrule
        \end{tabular}
      \end{sc}
    \end{small}
  \end{center}
  \vskip -0.1in
\end{table}

\section{Related Work}

\textbf{Point Cloud Self-Supervised Learning.} Inspired by the success of Masked Autoencoders in images \citep{he2022MAE}, reconstruction based point cloud SSL has largely adopted the 3D patch reconstruction framework established by PointBERT \citep{yu2021pointbert}. These methods typically employ a binary masking strategy on local patches, where specific regions of the point cloud are entirely removed from the input~\citep{pang2022pointMAE, chen2023pointgpt, liang2024pointmamba}. However, these frameworks lack inherent SE(3)-symmetry and their embeddings remain sensitive to global rotations and translations. RI-MAE pursues rotation invariance without equivariant architectures by PCA-aligning patches and reconstructing in latent space \citep{RI_MAE}. RIVQ-VAE learns a discretized, SE(3)-invariant patch representation using TFN~\citep{RIVQVAE}. All of these methods produce patch embeddings, requiring fine tuning with a task-specific head for downstream tasks. SIEVE instead uses equivariant networks and a spherical projection mask, producing node-level SE(3)-invariant embeddings that concatenate cleanly with raw coordinates as input to any downstream architecture, without fine-tuning the encoder.

\textbf{Equivariant Neural Networks.} To date, numerous equivariant architectures have been proposed ~\citep{vector-neuron, cen2024HEGNN, pelaez2024torchmdnet, aykent2025gotennet, 2021EPN}. However, many of these models have focused on molecular biology applications \citep{han2025equisurvey}; QM9 and MD17 molecular prediction datasets are some of the most popular benchmarks for equivariant neural networks~\citep{ramakrishnan2014QM9, chmiela2017MD17}. Molecules possess a significant SE(3)-invariant node feature, the atomic number, which distinguishes the domain from purely geometric data like point clouds. Consequently, the only existing work that combines SSL with SE(3)-equivariant networks is not directly applicable to point clouds due to this difference in domain \citep{an2025equimoleculeprediction}. To our knowledge, this is the first work to combine point cloud SSL with SE(3)-equivariant neural networks for general geometric learning.

\section{Conclusion}

In this work, we theoretically established that conventional masking strategies in geometric self-supervised learning induce a representational collapse when applied to SE(3)-equivariant architectures. Using the theory of vector bundles, we showed that isotropic or randomized masking functions which ignore innate geometric features of point clouds do not have enough capabilities to reconstruct such point clouds, because they cannot effectively reconstruct a wide range of classes of non-isomorphic vector bundles.

Based on these theoretical observations, we proposed SIEVE, a pioneering framework that bridges the gap between equivariance and SSL for point clouds. By utilizing a novel spherical projection masking strategy, SIEVE circumvented the topological obstructions inherent in isotropic masking, enabling robust geometric reconstruction. Our experiments empirically confirmed the representational collapse and demonstrated that SIEVE learns highly generalizable embeddings that outperform traditional descriptors in downstream tasks. To our knowledge, SIEVE is the first framework to combine self-supervised learning with SE(3)-equivariant networks for general point cloud geometry, and we hope it spurs further work at this intersection.

\begin{ack}
This research was supported by the ``Advanced GPU Utilization Support Program'' funded by the Government of the Republic of Korea (Ministry of Science and ICT).

This work was initiated while Jongann Lee was a paid intern at SolverX in the Republic of Korea and continued after the internship concluded. Following the internship, Jongann Lee was no longer employed by SolverX and received no monetary compensation from SolverX for his subsequent contributions. SolverX continued to provide in-kind research support through access to GPU computing resources and LLM-based tools.

Sun Woo Park thanks the Max Planck Institute for Mathematics for its generous support.

\textbf{Competing interests.}
Jongann Lee was previously a paid intern at SolverX in the Republic of Korea. SolverX subsequently continued to provide the in-kind research resources described above. The authors declare no other competing interests.
\end{ack}

\bibliographystyle{abbrvnat}
\bibliography{citations}


\appendix

\section{Mathematical Background} \label{appendix:mathematical-background}

We recommend interested readers to review \citep{HatcherAlgebraicTopology} for an introduction to algebraic topology (where definitions of homotopy, homotopy equivalence, homotopy groups, closed orientable manifolds, and fiber bundles can be found), and Section 1 of \citep{HatcherVectorBundle} for an introduction to the theory of vector bundles over topological spaces.

\subsection{Motivation for using vector bundles} \label{appendix:motivation}

We motivate the formulation of deep learning techniques as an approximator of global section of vector bundles by utilizing insights from the main results of Park et al~\citep{PCCW26}. 

\textbf{Universal approximation theorem}: 
The classical formulations of universal approximation theorem states that a feedforward neural network of arbitrary depth with non-polynomial activation function  can approximate any given continuous function $f: \mathbb{R}^m \to \mathbb{R}^n$. Such an approximation theorem ensures a neural network with arbitrary depth (or arbitrarily many weights and parameters) to find any auxiliary relation among collections of datasets, as long as they embed in Euclidean space $\mathbb{R}^m$. 

However, such a formulation of the universal approximation theorem does not necessarily imply that a neural network with arbitrary depth (or arbitrarily many weights and parameters) can find auxiliary relations among collections of datasets defined over manifolds $M$ which are not homeomorphic to $\mathbb{R}^m$. For example, given a manifold $M$ with open affine charts $\{U_\alpha\}_{\alpha \in \mathcal{A}}$, a neural network can approximate a continuous function $f_\alpha: U_\alpha \to \mathbb{R}^m$ for each $\alpha \in \mathcal{A}$, but one cannot guarantee that such a neural network will learn how to coherently glue such functions $f_\alpha: U_\alpha \to \mathbb{R}^m$. In terms of datasets associated to 3D point clouds, the functions $f_\alpha: U_\alpha \to \mathbb{R}^m$ correspond to measurements taken over 3D point clouds which constitute a part of (but not all) underlying geometric space.

\textbf{Using vector bundles}: By using the notion of vector bundles (see Definition \ref{def:vector_bundle} for its definition), we can reformulate the problem of gluing functions $f_\alpha$ as follows. Given a vector bundle $V \to M$, each function $f_\alpha: U_\alpha \to \mathbb{R}^m$ corresponds to local sections of the vector bundle restricted to the open affine chart $U_\alpha$. Here, the term ``local'' means that the function $f_\alpha$ is only well defined over the neighborhood $U_\alpha$, and it is not necessarily defined over $M$. One can construct a global section $f: M \to V$ by ensuring that the collection of functions $\{f_\alpha\}_{\alpha \in \mathcal{A}}$ satisfy the condition that given any two indices $\alpha, \beta \in \mathcal{A}$, the restrictions of functions $f_\alpha$ and $f_\beta$ over $U_\alpha \cap U_\beta$ are equal. By allowing such a condition to hold, one extends the function from $U_\alpha$ to $U_\alpha \cup U_\beta$. Iterating the procedure for all indices $\alpha \in \mathcal{A}$ allows us to construct the global section $f: M \to V$.

\textbf{Non-triviality of vector bundles}: We note that the global section need not be a function $f: M \to \mathbb{R}^m$. If a global section is given by a function $f: M \to \mathbb{R}^m$, then the vector bundle $V \to M$ is a trivial bundle, i.e. $V \cong M \times \mathbb{R}^m$. Otherwise, we say that the vector bundle $V \to M$ is non-trivial. Such non-triviality of vector bundle implies that there are obstructions in naively gluing local sections $f_\alpha: U_\alpha \to \mathbb{R}^m$. These obstructions can be detected by computing topological invariants of vector bundles, such as their homology or cohomology groups. Loosely speaking, a global section of a non-trivial vector bundle can be thought of as of datasets whose measurements depend crucially on their underlying geometric space (or an underlying manifold $M$). Meanwhile, a global section of a trivial vector bundle can be considered as of datasets whose measurements can be well approximated by classical deep neural networks without making significant reference to underlying geometric space of the datasets. In particular, this observation hints that even if the measurements of 3D point clouds defined over an underlying manifold $M$ take values in $\mathbb{R}^m$, we cannot guarantee that such measurements can be continuously extended to a function $f: M \to \mathbb{R}^m$. 

\textbf{Strength of deep learning algorithms}: We note that there are variants of universal approximation theorem which regards approximating functions between differentiable manifolds with neural networks of arbitrary width and depth, see for example the work of Kratsios and Papon~\citep{KP22}. On the other hand, deep learning algorithms of finite depth and width may not have enough capabilities to approximate global sections of any vector bundle $V \to M$ of a manifold. In light of this vein, the main result of Park et al~\citep{PCCW26} (see for example Theorem 1.2, Sections 2 and 3) gives a presheaf/copresheaf theoretic characterization of capabilities of message passing neural networks: they approximate global sections of presheaf (and copresheaf) of continuous functions induced from direct sums of skyscraper cosheaves (and respectively sheaves). The obstructions of these presheaves (and copresheaves) from being sheaves (and cosheaves, respectively) can be utilized to explain limitations of a wide range of classes of deep learning algorithms, such as non-unique gluing of local sections of such presheaves (and copresheaves), and adversarial attacks. 

\textbf{Deep learning algorithms and vector bundles}: Given that limitations of such deep learning algorithms originate from obstructions of a presheaf (and copresheaf) from being a sheaf (and respectively cosheaf), the paper hence calls for a need to design neural networks which are capable of approximating global sections of sheaves (or cosheaves) other than the aformentioned classes of presheaves and copresheaves. In particular, vector bundles over manifolds are examples of sheaves over manifolds. A neural network which has capabilities to approximate global sections of vector bundles over manifolds hence has potential to overcome limitations that other previously studied deep learning algorithms may have experienced. We can ask a valid question on what isomorphism classes of vector bundles $V \to M$ of a manifold a given neural network can approximate their global sections. The capability that a neural network can effectively analyze dataset dependent on its intrinsic geometry can be rephrased as the capability to approximate global sections of non-trivial isomorphism classes of vector bundles over $M$. In turn, it implies that the neural network has capabilities in detecting exotic underlying geometric features of a given dataset, and can perform on par with or beyond other neural network algorithms.

\subsection{Proof of Theorem~\ref{thm:main_theorem}} \label{appendix:proof-main_theorem}

We first recall the definition of tautological bundle.
\begin{definition}
  \label{def:tautological_bundle}
  Recall the homeomorphism of complex manifolds $S^2 \cong \mathbb{C}\mathbb{P}^1$, which is the Grassmannian of complex projective lines in $\mathbb{C}^2$. 
  The tautological bundle $\pi_t: T \to S^2$ is the vector bundle defined as follows: for each $x \in S^2$, the homeomorphism $S^2 \cong \mathbb{CP}^1$ identifies $x$ with a point that represents a complex projective line $\ell \subset \mathbb{C}^2$. The fiber $\pi_t^{-1}(x)$ is given by the 1-dimensional complex vector space $\ell$ itself. We note that $\pi_t: T \to S^2$ can also be interpreted as a 2-dimensional real vector bundle by using the fact that $\mathbb{C}$ is a 2-dimensional $\mathbb{R}$-vector space.
\end{definition}

The proof follows from applying Proposition 1.14 of Hatcher\citep{HatcherVectorBundle}. The isomorphism classes of $n$-dimensional real oriented vector bundles over $S^2$ is isomorphic to the fundamental group $\pi_1(SO(n))$. This group is isomorphic to $\pi_1(SO(3))$. To see this, we consider the short exact sequence of topological spaces obtained from fiber bundle of $S^n$,
  \begin{equation}
      SO(n) \to SO(n+1) \to S^n,
  \end{equation}
  from which the long exact sequence of homotopy groups can be obtained using Theorem 4.41 of \citep{HatcherAlgebraicTopology}:
  \begin{equation}
      \cdots \to \pi_2(S^n) \to \pi_1(SO(n)) \to \pi_1(SO(n+1)) \to \pi_1(S^n) \to \cdots.
  \end{equation}
  The triviality of higher homotopy groups $\pi_2(S^n) = \pi_1(S^n) = 0$ for $n \geq 3$, see in particular Corollary 4.9 of \citep{HatcherAlgebraicTopology}, implies that we have $\pi_1(SO(n)) \cong \pi_1(SO(n+1))$ for all $n \geq 3$. By induction we have $\pi_1(SO(n)) \cong \pi_1(SO(3))$ for all $n \geq 3$. But using the homeomorphism $SO(3) \simeq \mathbb{RP}^3$, we obtain
  \begin{equation}
      \pi_1(SO(3)) \cong \pi_1(\mathbb{RP}^3) \cong \mathbb{Z}/2\mathbb{Z}.
  \end{equation}
  Hence, every vector bundle over $S^2$ is either isomorphic to the trivial vector bundle $\mathbb{R}^n \times S^2 \to S^2$, or a non-trivial vector bundle of dimension $n$. We note that because $\pi_t: T \to S^2$ is a non-trivial real vector bundle of dimension 2. (As a complex line bundle, $L \otimes L$ gives the holomorphic tangent bundle of $S^2$.) Because $n \geq 3$, every non-trivial vector bundle of dimension $n$ is isomorphic to the vector bundle obtained by taking direct sum of fibers of $L$ with those of the trivial vector bundle $\mathbb{R}^{n-2} \times S^2 \to S^2$. This finishes the proof of the theorem.

\subsection{Additional elaboration and proof of Theorem~\ref{thm:constant_obstruction}} \label{appendix:proof-constant_obstruction}

We recall the following application of Proposition 1.4 to our setup from \citep{HatcherVectorBundle}.
\begin{proposition} \label{appendix:prop:trivial_bundle}
    Given a vector bundle $\pi: V_Y' \to Y'$ over a closed smooth manifold $Y' \subset \mathbb{R}^3$, there exists a vector bundle $E \to Y'$ such that $V_Y' \oplus E$ is the trivial vector bundle over $Y'$. 
\end{proposition}
The construction of the vector bundle $E$ is obtained by finding a finite open cover $\{U_i\}_{i=1}^J$ of $Y'$, over which the restriction of the vector bundle $V_Y' \to U_i$ is the trivial vector bundle $U_i \times \mathbb{R}^m \to U_i$ (often denoted as local trivialization of $V_Y'$). Using Urysohn's lemma, there exists a function $g_i: V_Y' \to \mathbb{R}^m$ which factors through the local trivialization $V_Y' \to U_i$. One can then define a map $g: V_Y' \to \mathbb{R}^{mJ} = \prod_{i=1}^J \mathbb{R}^m$ by defining $g := (g_1, g_2, \cdots, g_J)$. Then the map $(g, \pi): V_Y' \to \mathbb{R}^{mJ} \times Y'$ defines the isomorphism of vector bundles. We note that the vector bundle $E \to Y'$ is of dimension $(J-1)m$. Because $Y'$ is homotopy equivalent to a smooth orientable closed surface of genus $g$, there must be at least $2$ open covers $U_1, U_2$, over each of which one obtains a trivialization of vector bundle $V_Y' \to U_i$. 

In the case of constant masking function, we may consider the subspace $\mathbb{R} \subset \mathbb{R}^N$ spanned by the fixed choice of a vector $v \in \mathbb{R}^N$. Hence, constant masking function corresponds to replacing the vector bundle $V_X$ with $\mathbb{R} \times X$, because the range of the masking function lies in $\mathbb{R} \times X$ instead of $\mathbb{R}^N \times X$. This implies that if $\mathfrak{m}: X \to \mathbb{R}$ is a constant masking function, then the only possible way to embed a non-trivial vector bundle $V_Y' \to Y'$ into $Y' \times \mathbb{R} \to Y'$ is by establishing an isomorphism of vector bundles between $V_Y' \cong Y' \times \mathbb{R}$, a contradiction. This finishes the proof of Theorem~\ref{thm:constant_obstruction}.

We note that in the case where $\mathfrak{m}: X \to \mathbb{R}^N$ is a masking function that is not a section of a non-trivial vector bundle, the proof of the above proposition requires that the dimension of $E$ has to be at least $m$. Therefore, unless $N \geq 2m$, the above method cannot be utilized to ensure that the non-trivial vector bundle $V_Y' \to Y'$ can be regarded as a vector sub-bundle of a trivial vector bundle of dimension $N$. Even if one enlarges the parameter $N$ to be greater than $2m$, finding exactly what vector sub-bundle of the trivial bundle is the non-trivial bundle $V_Y'$ is a challenging task. This is because the construction of the map $(g, \pi)$ depends crucially on the choice of the open cover $\{U_i\}_{i=1}^J$ over which the non-trivial vector bundle $V_Y'$ admits local trivialization. It would be a very difficult task for the neural network to predict such geometric data regarding open covers $U_i$, especially in cases where even the auxiliary space $Y'$ parametrizing the point clouds is not understood correctly.

We also elaborate more on the relation between Theorem \ref{thm:constant_obstruction} and Figure \ref{figure:synthetic_reconstruction}. We thank an anonymous reviewer for pointing out the need to clarify the relation between these two results. Let $f:Y' \to V_Y'$ be a non-trivial global section. We note that $f$ shall be regarded as a restriction of a continuous function $F: \mathbb{R}^3 \to \mathbb{R}^{m_N}$ to $Y'$. Because $Y'$ is compact, Heine-Cantor theorem implies that $f$ is uniformly continuous. Hence, given any $\epsilon > 0$, there exists $\delta > 0$ such that for any $x, y \in Y'$ with $|x-y| < \delta$, we have $|f(x) - f(y)| < \epsilon$. Given that the learning objective of SIEVE is to optimize $L_1$ norm, we will denote by $| \cdot |$ the $L_1$ norm over $\mathbb{R}^3$ and $\mathbb{R}^{m_N}$, respectively.

Let $\Phi$ be the $\mathrm{SE}(3)$-equivariant neural network with constant masking function and scalar pipeline perturbed by Gaussian noise. Suppose $\Phi$ effectively approximates $f$. Then for every $x \in Y'$ we have $|\Phi(x) - f(x)| < \epsilon$. Fix a point $x \in Y'$. Using triangle inequality and uniform continuity of $f$, for every such $y \in Y'$ with $|x-y| < \delta$ we have
\begin{equation}
    \epsilon > |\Phi(x) - f(x)| \geq |\Phi(x) - \Phi(y)| - |\Phi(y)-f(y)| - |f(y) - f(x)| \geq |\Phi(x) - \Phi(y)| - 2\epsilon.
\end{equation}
Hence, we require that for every $y \in Y'$ such that $|x-y| < \delta$, we have $|\Phi(x) - \Phi(y)| < 3 \epsilon$, i.e. $\Phi$ is also uniformly continuous. Since the masking function is constant, there exists a vector $v \in \mathbb{R}^m$ for some $m > 0$ such that
\begin{equation}
    \Phi(x) = \mathrm{Decoder}(v, \mathrm{Encoder}([\mu_x, x])),
\end{equation}
where $\mu_x$ denotes perturbation by Gaussian noise $\mathcal{N}(0,I)$, and the appropriate projection maps for encoder and decoder functions are abbreviated. Here, given $a \in \mathbb{R}^{n_1}$ and $b \in \mathbb{R}^{n_2}$, the notation $[a,b] \in \mathbb{R}^{n_1 + n_2}$ denotes concatenation of vectors $a$ and $b$. Both encoder and decoder functions from GATr are continuous. Hence, there exists some $\delta^* > 0$ such that 
\begin{equation}
    |[\mu_x, x] - [\mu_y,y]| < \delta^* \Rightarrow |\Phi(x) - \Phi(y)| < 3 \epsilon.
\end{equation}
From our assumption, we had $|x-y| < \delta$. Hence, by the definition of $L_1$ norm and triangle inequality, we have
\begin{equation}
    \delta^* > |[\mu_x,x] - [\mu_y,y]| = |[\mu_x-\mu_y, x-y]| = |\mu_x - \mu_y| + |x-y|.
\end{equation}
This implies that for any $y \in Y'$ such that $|x-y| < \delta$, we have $|\mu_x - \mu_y| < \delta^* - \delta$. However, because $\mu_x$ and $\mu_y$ are identically independently distributed, and there are arbitrarily many points lying in an open neighborhood of $x$ of radius $\delta$, the inequality $|\mu_x - \mu_y| < \delta^* - \delta$ cannot hold uniformly for all such $y \in Y'$ almost surely, unless $\delta^*$ is arbitrarily large. Hence, with such $\delta^*$ to be chosen to be arbitrarily large, in order for the inequality $|\Phi(x) - \Phi(y)| < 3 \epsilon$ to be satisfied, we require that there exists some arbitrarily small $\epsilon'' > 0$ such that either $\sup_{x \in Y'}|\mathrm{Encoder}([\mu_x,x])| < \epsilon''$, or $\sup_{x \in Y'}|\Phi(x)| < \epsilon''$. Suppose the former holds. (We note that the latter case implies that $f$ is a trivial function, contradicting our assumption on $f$). This implies that the domain of the decoder function (the last component of $\Phi$) is contained in a compact subset corresponding to a ball of radius $\epsilon'' > 0$ from the constant vector $v \in \mathbb{R}^m$ that we chose earlier. Within this domain, the decoder function is uniformly continuous, hence one can choose $\epsilon > 0$ to be arbitrarily small so that the image of $\Phi$ is mapped to a ball of arbitrarily small radius centered at $\Phi(v)$. This explain the empirical collapse in Figure \ref{figure:synthetic_reconstruction}.

In relation to Theorem~\ref{thm:constant_obstruction}, the above argument shows that if $\Phi$ were to approximate the global section $f: Y' \to V_Y'$ effectively, then $f$ would have been a constant function $\Phi(v)$ for some fixed vector $v$ with arbitrarily small perturbations. We can then proceed as in the proof of Theorem~\ref{thm:constant_obstruction} stated above to conclude that the only global sections of vector bundles such $\Phi$ can approximate (almost surely) are rank $1$ trivial bundle of form $\mathbb{R} \times Y' \to Y'$.

\subsection{Proof of Theorem~\ref{thm:SIEVE_main}} \label{appendix:proof_SIEVE_main}
The theorem follows from considering the reformulation of the architectural components of SIEVE into the following steps as follows.
\begin{enumerate}
    \item Using 3D point clouds, consider an auxiliary (possibly unknown) $n$-dimensional vector bundle $V_Y \to Y$' of some 3-dimensional closed orientable manifold $Y$' homeomorphic to $X \times I$ for some closed orientable surface $X \subset \mathbb{R}^3$ of genus $g$. We regard the $n$-dimensional measurements over 3D point clouds correspond as a global section $f: Y' \to V_Y$ of the $n$-dimensional vector bundle over $Y'$.
    \item \textbf{(Encoder):} Find an $m$-dimensional vector bundle $V_Y$' $\to Y$' with $m \geq 2n+5$ such that the fiber of $V_Y$ injects into a vector subspace of the fiber of $V_Y$' at every $y \in Y$'.
    \item Consider an auxiliary $m$-dimensional vector bundle $V_X \to X$ such that $V_X$ and $V_Y'$ are equivalent vector bundles with respect to homotopy equivalence $X \times I \to X$.
    \item \textbf{(Spherical Masking):} Using the spherical projection $\pi_2: X \to S^2$ which is the continuous map, replace the $m$-dimensional vector bundle $V_X \to X$ with the pullback of some $m$-dimensional vector bundle $V \to S^2$. This replacement operator is well defined if $V_X$ is isomorphic to the pullback bundle $\pi_2^* V$ as vector bundles over $X$, which is why the statement of the theorem makes technical assumption on isomorphism class of vector bundles that SIEVE can approximate their global sections. By definition of spherical masking, $V$ contains the dimension $2$ tautological bundle $\pi_{t}: T \to S^2$ as its sub-vector bundle. Using classification of vector bundles over $S^2$, the vector bundle $V \to S^2$ must be isomorphic to either the trivial vector bundle or the unique non-trivial vector bundle of dimension $m$, whose fibers given by a summand of form $\mathbb{R}^{m-2} \oplus \pi_t^{-1}(x)$ for every $x \in S^2$. 
    \item The spherical projection transforms the $n$-dimensional measurements over 3D point clouds via the composition of maps
    \begin{equation}
        f^*: Y' \to V_Y \hookrightarrow V_Y' \cong V_X \cong \pi_2^* V \to \pi_2^* T.
    \end{equation}
    Here, the first map is our global section $f: Y' \to V_Y$. The second map is the inclusion of fibers of $V_Y$ into $V_Y'$ (i.e. the encoder). The third map originates from equivalence of vector bundles between $X$ and $Y'$. The fourth and fifth map originates from replacing the vector bundle $V_X$ with the pullback bundle of $V \to S^2$ using the spherical masking function. Note that the fifth map originates from the projecting the vector bundle $\pi_2^*V$ to its vector sub-bundle $\pi_2^* T$. (The previous step implies that $\pi_2^*V \cong \mathbb{R}^{m-2} \oplus \pi_2^* T$, so the fifth projection map can be explicitly define by mapping $\mathbb{R}^{m-2}$ to $\{0\}$).
    \item \textbf{(Decoder):} Reconstruct the $n$-dimensional vector bundle $V_Y \to Y$' as a closed orientable $n+2$ dimensional submanifold of $\mathbb{R}^{m+3}$, and approximate the global section $f: Y' \to V_Y$ using the spherical projection of measurements of 3D point cloud $f^*: Y' \to \pi_2^* T$. 
\end{enumerate}
The fact that spherical projection masking corresponds to replacing the vector bundle with the pullback of the non-trivial vector bundle over $S^2$ can be shown from the following reasoning. Using the definition of pullback of vector bundles and tautological bundles, the feature vectors obtained from the spherical masking function $\mathrm{SphericalMask}$ lie in the span of a linear subspace of $\mathbb{C}^2$ that passes through the mean $\overline{\mathbf{x}}$. Up to linear change of coordinates that sends $\overline{\mathbf{x}}$ to the point of origin $0 \in \mathbb{C}^1$, we obtain that the spherical masking function $\mathrm{SphericalMask}$ corresponds to defining a global section $g: X \to \pi_2^*T$, where for each $x \in X$, we have $g(x) = \mathrm{SphericalMask}(x)$. For general vector bundles $V_X \to X$, we can hence use $\mathrm{SphericalMask}$ to replace them with pullback of vector bundle $V \to S^2$ which contains $T \to S^2$ as a vector sub-bundle.

The requirement for $m \geq 2n + 5$ need not be satisfied, but we explain the motivation behind such a choice. By construction, we may assume that the vector bundle $V_Y \to Y$' is a smooth compact orientable manifold of dimension $n+2$. By Haefliger-Hirsch, C.T.C. Wall, and Ghanwat-Pancholi’s results, every such manifold embed into $\mathbb{R}^{2(n+2)-1} = \mathbb{R}^{2n+3}.$ \citep{haefliger1963} \citep{wall1965all} \citep{ghanwat2020embeddings} Using the fact that $V$ contains fibers isomorphic to the Euclidean space $\mathbb{R}^{h}$ of dimension at least $m-2$, the condition on $m$ ensures to identify the desired vector bundle $V_Y$ as a smoothly embedded submanifold of the vector bundle $V \to S^2$. We note that the reconstruction of the desired vector bundle $V_Y \to Y$' can be obtained from applying the universal approximation theorem for approximating the desired continuous function $F:\mathbb{R}^{m+3} \to \mathbb{R}^{m+3}$ whose image is homeomorphic to the vector bundle $V_Y$. 

The approximation of the global section $f: Y' \to V_Y$ using $f^*: Y' \to \pi_2^*T$ follows from using universal approximation theorem for $\mathbb{R}^m$. Without loss of generality, we assume $V_Y$ is a non-trivial vector bundle which are isomorphic to pullback of a non-trivial vector bundle of $S^2$. The decoder satisfying conditions of universal approximation theorem for approximating continuous functions $f: \mathbb{R}^m \to \mathbb{R}^n$ can be used to approximate global sections of trivial vector bundle $g: Y' \to \mathbb{R}^{n-2}$ by approximating local restrictions of $g$ to an open affine chart $\{U_\alpha\}_{\alpha \in \mathcal{A}}$ of $Y'$ and gluing such local restrictions of $g$ without any obstructions. Because the vector bundle $V_Y$ is isomorphic to $\mathbb{R}^{n-2} \oplus \pi_2^* T$, we can use the product of functions $(g,f^*): Y' \to \mathbb{R}^{n-2} \oplus \pi_2^*T \cong V_Y$ to approximate any global section of $V_Y$. 

\subsection{Theoretical Limitations} \label{appendix:limitations}

Our theoretical results suggest that SIEVE is capable of approximating global sections of a wide range of classes of non-trivial vector bundles over manifolds. Nevertheless, as similar to other state-of-the-art techniques, SIEVE has some theoretical limitations, in that it does not guarantee outperformance in analyzing datasets which are global sections of vector bundles which do not contain pullback of tautological bundles as vector sub-bundle. For instance, there may be many non-trivial isomorphism classes of vector bundles over manifolds which do not originate from pullback of non-trivial vector bundles over $S^2$. Further studies would be required to devise new neural networks which are capable of approximating such global sections of nontrivial vector bundles.

\section{Implementation Details} \label{appendix:implementation-details}

\subsection{Pre-Training} \label{appendix:pretrain-details}

\textbf{Model Parameters: }The backbone of pre-trained SIEVE is GATr, a model based on multivector representations \citep{GATr2023}. We embed the position of each node as a trivector inside the multivector. 

The encoder comprises a 9-layer GATr with 16 multivector and 64 scalar channels. The input is 1 multivector and 1 scalar channel and the output is 1 multivector and 32 scalar channel. The 32-dimensional scalar output constitutes the SIEVE embedding $s_{code}$. The encoder has 949,825 parameters.

The decoder is a 3-layer GATr with 16 multivector and 64 scalar channels. The input is 1 multivector and 1 scalar channel and the output is a single-channel multivector and scalar output for coordinate reconstruction. The decoder has 318,690 parameters.

Note that for variants of SIEVE trained with ShapeNet~\citep{shapenet2015}, the model parameters from Table \ref{table:mask_ablation_hyperparam} are used.

\textbf{Training Method:} The model is trained for 10 epochs using the AdamW optimizer with a batch size of 200. To accelarate training, bf16 quantization is used. We employ a cosine learning rate schedule with a warm-up phase, reaching a maximum learning rate of $3 \times 10^{-4}$. To stabilize training, we apply gradient clipping with a maximum norm of 10. We train on a single Nvidia B200 GPU over 4 days. Memory consumption with a batch size of 200 is 154 GB.

\textbf{Additional Techniques:} Three distinct pre-training strategies have been tested on the smaller ShapeNet dataset~\citep{shapenet2015}.

The first strategy utilizes simple position reconstruction as outlined in the main text, with pre-training performed on a volumetric ShapeNet dataset. The standard deviation of the spherical projection mask noise is $\sigma = 0.02$.

The second strategy is to reconstruct the normal vector and the curvature in addition to the position of the node. The normal vector is the output multivector's vector, and the curvature is the output scalar. The reconstruction loss for this strategy is 
\begin{equation} 
    \mathcal L = L1(\mathrm{PredPos, Pos}) + 10 \cdot (1 - \cos(\mathrm{PredNorm} \cdot \mathrm{Norm}) + 5 \cdot L1(\mathrm{PredCurv, Curv}).
\end{equation}
We use the surface ShapeNet dataset for this strategy, as neither curvature and normal vectors can be defined in the interior.

The third strategy involves the application of a nearest-neighbor attention mask within the decoder. To ensure the encoder extracts global embeddings, the decoder is forced to reconstruct the nodes without local information by masking $70\%$ of the nearest nodes during self-attention. This strategy is implemented on a volumetric ShapeNet dataset.

As discussed in Appendix \ref{appendix:spherical-projection-collapse} and Table \ref{table:RNA_Ablation}, the third strategy of using a nearest-neighbor attention mask yielded the best results. Therefore, the final OpenShape~\citep{2023openshape} pretrained SIEVE utilizes a nearest neighbor attention mask with ratio $0.7$ along with a spherical projection mask.

\subsection{Synthetic Data Experiment} \label{mask-ablation-experiment-details}

\textbf{Model Parameters:} The mask ablation study (Section \ref{subsec:MaskAblation}) evaluates two different equivariant backbone architectures: GATr \citep{GATr2023} and SE(3)-Transformer \citep{fuchs2020se3transformers}. As detailed in Table \ref{table:mask_ablation_hyperparam}, both models utilize matching hidden channels and layer numbers to ensure a fair comparison. The positions are embedded as trivectors for GATr and as type-1 vectors for the SE(3)-Transformer. For the SE(3)-Transformer, we generate edges between the nodes using $k$-nearest neighbor with $k=50$.

\begin{table}[ht]
  \caption{Model Hyperparameters for the Mask Ablation Study}
  \label{table:mask_ablation_hyperparam}
  \begin{center}
    \begin{small}
      \begin{sc}
        \begin{tabular}{llcc}
          \toprule
          Module & Hyperparameter & GATr & SE(3)-Transformer \\
          \midrule
          \multirow{4}{*}{Encoder}
            & Input Channel (s,v)       & (1,1)  & (1,1)  \\
            & Hidden Channel (s,v)      & (32,8) & (32,8) \\
            & Output Channel (s,v)      & (16,1) & (16,1) \\
            & Number of Attention Heads & 4      & 1  \\
            & Number of Layers          & 6      & 6  \\
          \cmidrule{1-4}
          \multirow{4}{*}{Decoder}
            & Hidden Channel (s,v)      & (32,8) & (32,8) \\
            & Output Channel (s,v)      & (1,1)  & (1,1)  \\
            & Number of Attention Heads & 4  & 1  \\
            & Number of Layers          & 2  & 2  \\
          \bottomrule
        \end{tabular}
      \end{sc}
    \end{small}
  \end{center}
  \vskip -0.1in
\end{table}

\textbf{Mask Details:} The constant mask is a single vector or multivector token that is applied uniformly over all of the nodes. This token is trainable. The noise mask is a uniform sample of points on an ellipsoid surface with semi-axes $(2,1,0.5)$. The ellipsoid is centered at each point cloud's centroid, and the orientation is determined by PCA of the original point cloud: x-axis for the largest variance direction, y-axis for the middle, and z-axis for the smallest. This ensures that the noise mask is also an equivariant operator. The spherical projection mask is as described in Section \ref{subsec:Spherical_Proj_Mask}, but no noise is added to the projected position. 

\textbf{Training Method:} Both models are trained with a simple $\mathcal L1$ loss on predicted node positions. The models are trained over 200 epochs, with a batch size of 4. We employ a cosine learning rate decay that reaches a maximum learning rate of $3 \times 10^{-4}$ after a $10\%$ warm up period. To ensure numerical stability, the gradients are clipped at a threshold of 10.0. No weight decay is applied.

\subsection{Downstream Tasks} \label{appendix:downstream-task-details}

Each downstream task utilizes a domain specific architecture. For the 3D-EPN point cloud completion task, we utilize PCN \citep{yuan2018PCN}; for the point cloud registration task on ModelNet40, we utilize DCP \citep{wang2019dcp}. All downstream models use the hyperparameters specified in their original publications and official implementations. The only modification made is to the input channel to accommodate the additional 32-dimension $s_{code}$ embeddings. The SIEVE embeddings are concatenated along with the position as the input. The SIEVE encoder is frozen when used in a downstream task; we do not fine-tune it. The RNA segmentation task and the geometric registration task use the OpenShape pretrained encoder, and the 3D shape completion task uses the ShapeNet pretrained encoder with the "simple noise" setting.

The added downstream task training time is a single forward pass through the frozen SIEVE encoder. On the RNA segmentation task with PointNet++, memory increases from 2134 MiB to 2166 MiB, and per-epoch time from 92s to 123s (+31s). On 3D-EPN with PCN, per-epoch time increases from 4 min 56s to 5 min 43 sec (+47s). The absolute overhead is comparable in both cases, so its relative share falls as the downstream backbone becomes more expensive: 34\% for PointNet++ against 16\% for PCN.

\section{Ablation Study} \label{appendix:ablation}

\subsection{Reconstruction of Point Clouds with Identical Topology} \label{appendix:deepjeb}

To demonstrate that the topological obstruction is the cause of the representational collapse, we train SIEVE to reconstruct a point cloud dataset with identical topology. We use DeepJEB, which contains 2138 meshes of jet engine brackets \citep{hong2025deepjeb}. These engine brackets all have two holes at the top and 4 holes at the bottom; therefore all of the meshes have the same topology. 

\begin{figure}[h]
  \centering
  \begin{subfigure}[b]{0.33\textwidth}
    \centering
    \includegraphics[width=\textwidth]{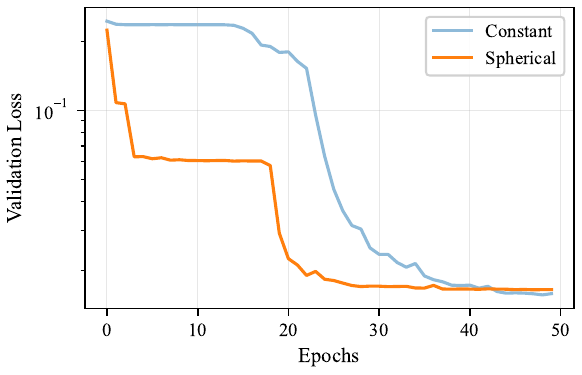}
    \caption{Reconstruction Validation Loss }
  \end{subfigure}
  \hfill
  \begin{subfigure}[b]{0.66\textwidth}
    \centering
    \includegraphics[width=\textwidth]{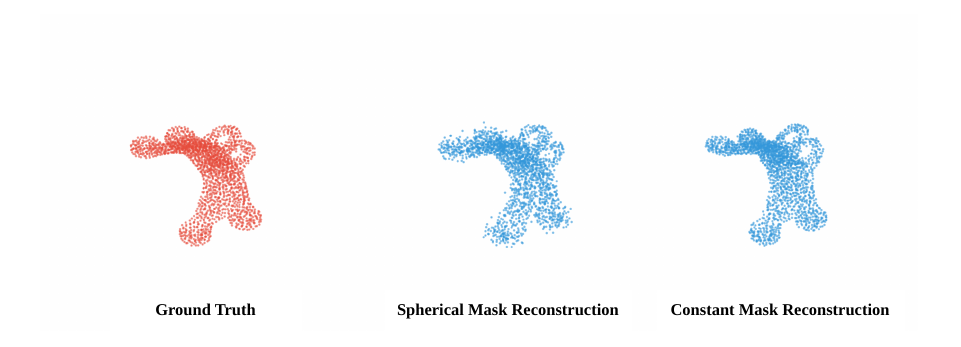}
    \caption{Reconstruction Result}
  \end{subfigure}
  \caption{\textbf{SIEVE on DeepJEB.} DeepJEB is a point cloud dataset that contains objects with identical topology. Therefore, both the constant mask SIEVE and the spherical mask SIEVE is able to do near-perfect reconstruction.}
  \label{fig:DeepJEB_reconstruction}
\end{figure}

The results (Figure \ref{fig:DeepJEB_reconstruction}) show that both constant mask SIEVE and spherical mask SIEVE are able to reconstruct point clouds in the DeepJEB dataset. Hence we confirm that it is the topological obstruction that is causing the representational collapse in Section \ref{subsec:MaskAblation}, not an inherent architectural limitation.

\subsection{Scalar Input of Encoder} \label{appendix:scalar-enc-input}

Our choice to use Gaussian noise as the scalar input of the equivariant encoder is based on its capability and ease of use. This subsection is an ablation study on the scalar input of the encoder to justify this selection.

We compare noise to two other scalar encoder inputs: a constant array of ones $\mathbf 1 = [1, 1, \cdots, 1]^T$ and a handpicked invariant scalar feature of the point cloud, for which we chose surface variance (SV). Given a node $\mathbf x_i$, and its 16 closest neighbors $\mathcal N(\mathbf x_i)$, the covariance matrix of the 16 neighbors' relative position $\mathbf p_{ij} = \mathbf x_j - \mathbf x_i, \mathbf x_j \in \mathcal N(\mathbf x_i)$ is defined as $K$. Denoting the three eigenvalue of $K$ as $\lambda_1\geq \lambda_2 \geq \lambda_3 $, sv is defined as $\mathrm{sv} = \lambda_1/(\lambda_1 + \lambda_2 + \lambda_3)$. 

Three distinct SIEVE models were trained with different encoder inputs. The ModelNet40 dataset was used to pre-train all of the models, and the backbone model for SIEVE was GATr. Training was conducted over 100 epochs with a batch size of 32, using a cosine learning rate decay with a maximum learning rate of $3 \times 10^{-4}$.

\begin{figure}[h]
  \vskip -0.1in
  \begin{center}
    \centerline{\includegraphics[width=\columnwidth]{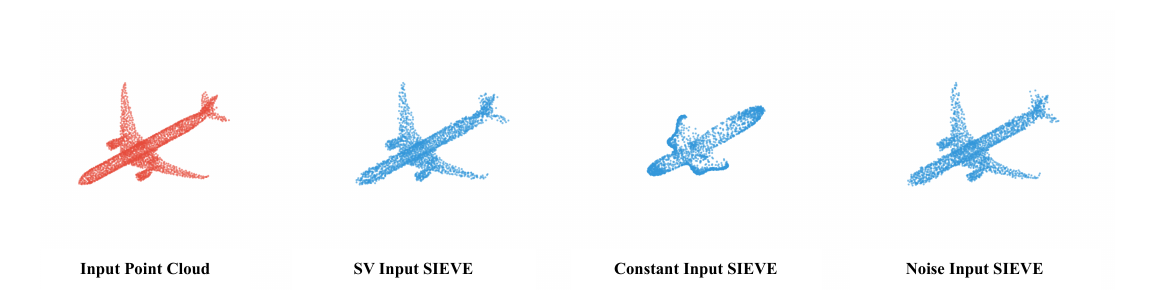}}
    \caption{
        \textbf{Input Noise Ablation.} We train SIEVE on ModelNet40 with three different input: local point cloud feature sv, a constant array of ones, and noise. Either noise or sv input enables SIEVE to reconstruct the input point cloud. SIEVE with constant input shows significantly worse performance when reconstructing the input point cloud.
    }
    \label{figure:input_ablation}
  \end{center}
  \vskip -0.1in
\end{figure}

Figure \ref{figure:input_ablation} showcases the reconstruction results of SIEVE with different encoder inputs. Both sv and noise are shown to be valid inputs; SIEVE is capable of reconstructing the input point cloud nearly perfectly with either. The constant array input is shown to be suboptimal. While it does not yield a complete collapse, it demonstrates difficulties in reconstructing the geometric details of the point cloud. 

We choose noise over sv because it does not require pre-processing of the input point cloud. Relying on handpicked SE(3)-invariant point cloud features increases the difficulty of implementing SIEVE across a wide range of datasets.

\begin{figure}[h]
  \vskip -0.1in
  \begin{center}
    \centerline{\includegraphics[width=0.6\columnwidth]{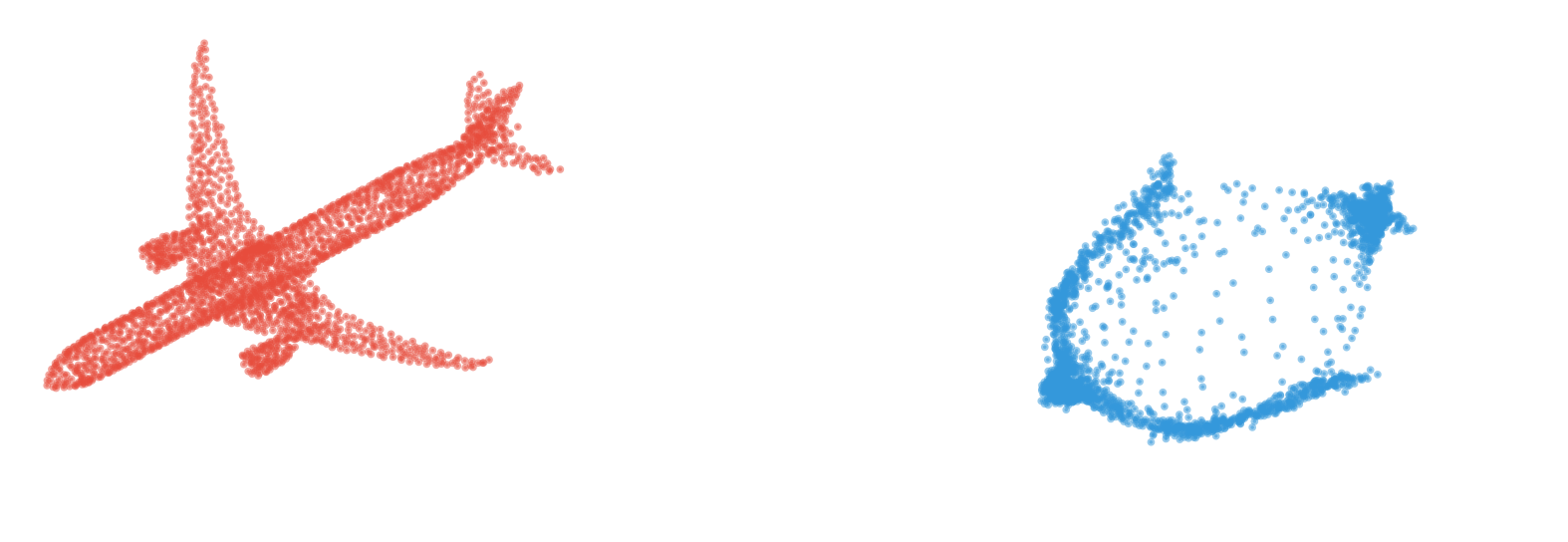}}
    \caption{
        \textbf{Noise Input Trained SIEVE with SV as Input.} The above figure demonstrates the result \emph{(blue, right)} of reconstructing the input point cloud \emph{(red, left)} using a noise-input pre-trained SIEVE with SV as scalar input. The distorted reconstruction implies that the encoder does not simply ignore the scalar noise input.
    }
    \label{figure:noise_trained_sdf_input_reconstruction}
  \end{center}
  \vskip -0.1in
\end{figure}

It is interesting to note that the use of noise as input does not lead to the scalar input being completely ignored. Figure \ref{figure:noise_trained_sdf_input_reconstruction} showcases the result when SIEVE trained with noise as input receives sv as input instead. The reconstruction is distorted, which would not be the case if the scalar input was simply ignored by the model. We therefore conclude that the encoder utilizes the specific distribution of the scalar input to construct the embedding, further highlighting the importance of the input signal even when it is stochastic.

The use of noise as an encoder input does not lead to SIEVE simply memorizing a specific noise realization for a given point cloud. Table \ref{table:scalar_input_seed} provides the results when using different random seeds to generate the noise input for the encoder. While each seed yields a unique input noise for the encoder, these differences have almost no effect on SIEVE's ability to reconstruct the point cloud. The three different seeds yield nearly identical validation losses, demonstrating that the model can consistently generate embeddings over various samples of noise.

\begin{table}[ht]
  \caption{Reconstruction Validation Loss $(\mathcal L1)$ of SIEVE with different random seeds on the ModelNet40 dataset}
  \label{table:scalar_input_seed}
  \begin{center}
    \begin{small}
      \begin{sc}
        \begin{tabular}{lccc}
          \toprule
               & seed 42 & seed 67 & seed 721 \\
          \midrule
          Validation Loss   & $1.004 \times 10^{-2}$  & $1.002 \times 10^{-2}$ & $0.9981 \times 10^{-2}$   \\
          \bottomrule
        \end{tabular}
      \end{sc}
    \end{small}
  \end{center}
  \vskip -0.1in
\end{table}

\subsection{SIEVE Embeddings do not Collapse to the Radius} \label{appendix:spherical-projection-collapse}

\begin{figure}[h]
  \vskip -0.1in
  \begin{center}
    \centerline{\includegraphics[width=\columnwidth]{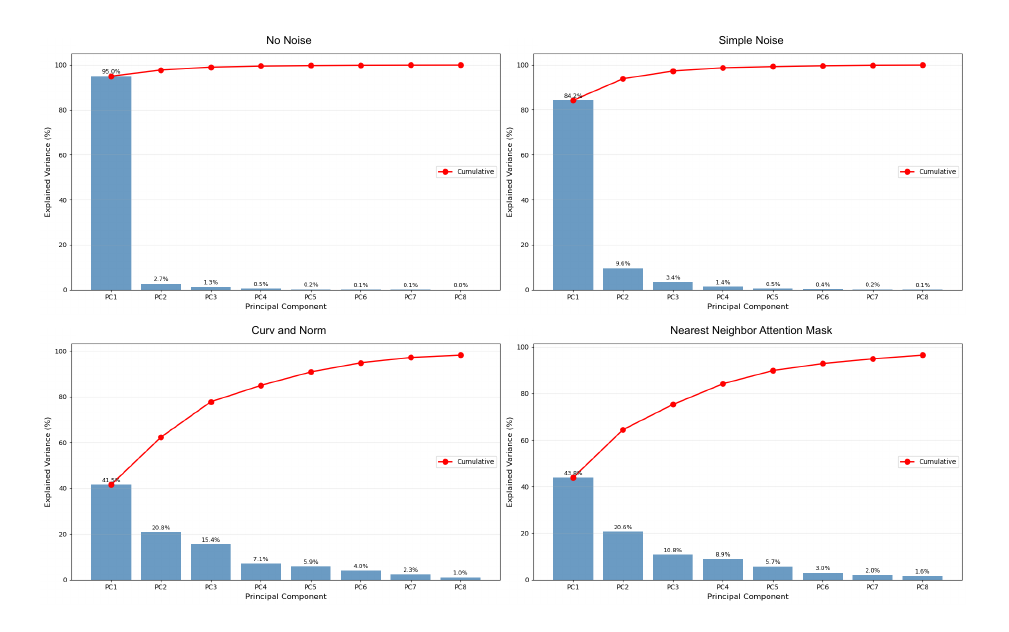}}
    \caption{
        \textbf{SIEVE Embedding PCA Analysis.} We perform PCA on the SIEVE embedding $s_{code}$ and plot the percentage of each eigenvalue $\lambda_i$ over the sum of all the eigenvalues. In addition, the cumulative percentage is also plotted. From the top left, we plot the analysis results for SIEVE with no noise, `simple noise' strategy, `curv and norm' strategy, and `nearest neighbor attention mask' strategy. }
    \label{figure:SIEVE_PCA}
  \end{center}
  \vskip -0.1in
\end{figure}

In this subsection, we rigorously analyze the embeddings to demonstrate that they do not collapse into trivial representations of the radius $r = \| \mathbf x_i - \overline{\mathbf x}\|$. We compare SIEVE using a simple spherical projection mask without any noise to a variant with noise added to the spherical projection. We refer to this variant as `simple noise'.

Additionally, we compare two further variations of SIEVE. The first reconstructs the normal vector and the curvature in addition to the position. We refer to this variant as `curv and norm'. The second applies an additional self-attention mask to the decoder, where for every node the nearest $70 \%$ of nodes are masked during self-attention. We refer to this variant as `nearest neighbor attention mask'. The models are pre-trained with the settings outlined in Appendix \ref{appendix:pretrain-details}.

We first analyze the embeddings using principal component analysis (PCA). Given an input point cloud and its SIEVE embedding $s_{code} \in \mathbb R^{N_p \times 16}$, the covariance of the embedding is $K = \frac{1}{N_p-1}[s_{code} - \mathrm{mean}(s_{code})]^T [s_{code} - \mathrm{mean}(s_{code})] \in \mathbb R^{16 \times 16}$. Since $K$ is a positive semi-definite matrix, all the eigenvalues are non-negative. We sort the eigenvalues in decreasing order $\lambda_1 \geq \lambda_2 \geq \cdots \geq \lambda_{16}$, and calculate the ratio of each eigenvalue relative to the total variance$\frac{\lambda_i}{\sum_{i=1}^{16} \lambda_i}$.

PCA analysis (Figure \ref{figure:SIEVE_PCA}) showcases that not adding any noise to the spherical projection mask yields a dimensional collapse, with $95\%$ of the variation contained in the first principal component. The addition of a simple Gaussian noise prevents this collapse, decreasing the first principal component's variation percentage to $84.2\%$, and increasing the second principal component's variation percentage to $9.6 \%$. Furthermore, using more difficult reconstruction strategies such as `curv and norm' and `nearest neighbor attention mask' mitigates the dimensional collapse even further, with both strategies showing variation being spread out over multiple principal components.

Next, we directly compare the learned embeddings to the radial distance $r = \| \mathbf x_i - \overline{\mathbf x}\|$. We calculate the Pearson correlation coefficient between each of the 16 channels of $s_{code}$ and the radius $r$ for a representative point cloud sample. 

\begin{figure}[h]
  \vskip -0.1in
  \begin{center}
    \centerline{\includegraphics[width=\linewidth]{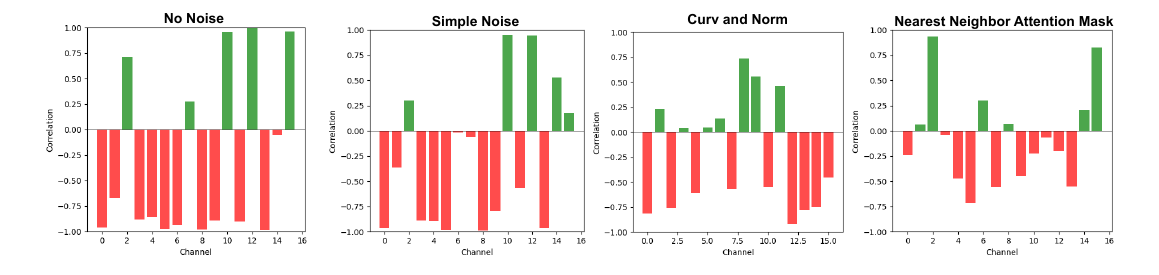}}
    \caption{
        \textbf{Embedding Radius Correlation.} This plot illustrates the correlation between each channel of the SIEVE embedding and the radius $r = \| \mathbf x_i - \overline{\mathbf x}\|$.
        }
    \label{figure:SIEVE_radius_corr}
  \end{center}
  \vskip -0.2in
\end{figure}

The correlation analysis (Figure \ref{figure:SIEVE_radius_corr}) demonstrates that when no noise is added to the masking process, the embeddings collapse and correlate heavily with the radius $r$. In this configuration, 12 out of 16 channels exhibit an absolute correlation value above 0.75. The addition of simple noise reduces this count to 9, indicating that even basic noise helps prevent a complete dimensional collapse to the radial component. The more sophisticated reconstruction strategies further decouple the embeddings from the radius; notably, the `nearest neighbor attention mask' variant results in only 2 channels possessing an absolute correlation value above 0.75.

Finally, we apply each strategy's embedding to the RNA segmentation downstream task with PointNet++ to evaluate the marginal utility of each variant compared to a raw radial feature. The results are in table \ref{table:RNA_Ablation}.

\begin{table}[ht]
  \caption{Test accuracies(\%) on the RNA segmentation task with different SIEVE strategies}
  \label{table:RNA_Ablation}
  \begin{center}
    \begin{small}
      \begin{sc}
        \begin{tabular}{cccccc}
          \toprule
          $xyz$ only & radius $r$ &No Noise & Simple Noise & Curv and Norm & NN Attention Mask \\
          \midrule
          71.55      & 75.54      & 76.08   &   75.94      &    78.05      & 84.99  \\
          \bottomrule
        \end{tabular}
      \end{sc}
    \end{small}
  \end{center}
  \vskip -0.1in
\end{table}

The SIEVE embedding pre-trained without noise yields results nearly identical to those using the radius $r$ as a feature, confirming the dimensional collapse identified in our PCA analysis. While the addition of simple noise disrupts this radial dependency, it initially results in a marginal performance decrease. In contrast, the more sophisticated strategies yield significantly higher accuracies. This indicates that these strategies have learned richer features beyond the simple radius $r$.

\subsection{SIEVE Encoder Needs to be Trained} \label{appendix:untrained-SIEVE}

Since SIEVE uses SE(3)-equivariant neural networks, even a randomly initialized SIEVE encoder provides an SE(3)-invariant embedding to the downstream task model. To isolate the contribution of the pretraining, we compare the downstream task performance of a randomly initialized SIEVE encoder against a pre-trained one. The setup reproduces Table \ref{table:RNA_Task_Table} exactly: same encoder architecture, $s_{code}$ integration and hyperparameters. The only difference is the weights loaded into the SIEVE encoder. In both arms, $s_{code}$ is passed to the downstream network exactly as the encoder emits it, with no rescaling or normalization. The results are below.

\begin{table}[h]
    \caption{Comparison of untrained SIEVE and pre-trained SIEVE on the RNA segmentation task (\%)}
    \label{table:RNA_Untrained_SIEVE}
    \begin{center}
        \begin{small}
            \begin{tabular}{lccc}
                \toprule
                 & PointNet++ & DeltaNet & DiffusionNet \\
                 \midrule
                 $xyz$ as feature & 71.55 & 87.21 & 81.20 \\
                 random init SIEVE & 67.47 & 77.59 & 82.52 \\
                 pre-trained SIEVE & 86.28 & 89.58 & 89.30 \\
                 \bottomrule
            \end{tabular}            
        \end{small}
    \end{center}

    \label{tab:placeholder}
\end{table}

Adding a randomly initialized SIEVE encoder to the downstream task does not meaningfully improve the performance. DiffusionNet’s accuracy increases marginally (+1.32), but both DeltaNet (-9.62) and PointNet++ (-4.08) suffer performance loss with a randomly initialized SIEVE encoder. Across all three backbones, neither the equivariant architecture nor the additional input width accounts for the gain in Table \ref{table:RNA_Task_Table}; those gains are attributable to the pretraining itself.

\subsection{SIEVE Embedding Injection Location} \label{appendix:injection-location}

The SIEVE encoder is frozen downstream, and every downstream task uses the original published hyperparameters; the only modification is widening the input layer. The choice that does matter is where the embedding is injected. We have done an ablation study on the geometric registration task where the ShapeNet pre-trained SIEVE embeddings are injected at different locations. 

\begin{table}[ht]
    \caption{$s_{code}$ injection location comparison on the geometric registration task}
    \label{table:injection-location-ablation}
    \begin{center}
        \begin{small}
            \begin{tabular}{lcccccc}
                \toprule
                $s_{code}$ injection location & MSE($R$) & RMSE($R$) & MAE($R$) & MSE($t$) & RMSE($t$) & MAE($t$) \\
                \midrule
                \textbf{Before DGCNN (chosen)} & 7.05 & 2.65 & 1.39 & 0.000004 & 0.0020 & 0.00092 \\
                DGCNN Intermediate & 7.34 & 2.71 & 1.55 & 0.000006 & 0.0025 & 0.0015 \\
                Before Transformer & 12.43 & 3.53 & 2.01 & 0.000029 & 0.0054 & 0.0036 \\
                Before SVD         & 24.59 & 4.96 & 3.12 & 0.000452 & 0.021 & 0.015 \\
                \bottomrule
            \end{tabular}
        \end{small}
    \end{center}
\end{table}

We have found that concatenating at the start yields the best results. Therefore, all of the downstream tasks concatenate $s_{code}$ at the downstream network input.



\end{document}